\RequirePackage{fix-cm}
\documentclass[smallextended]{svjour3}       % onecolumn (second format)
\smartqed  % flush right qed marks, e.g. at end of proof
\usepackage{graphicx}
\usepackage{amsmath,multirow,bm,amssymb,psfrag,algorithm,subfigure,color,mdframed,wasysym,subeqnarray,multicol}
\usepackage{epstopdf}
\usepackage[round]{natbib}
 \journalname{Data Mining and Knowledge Discovery}
\begin{document}

\title{Gaussian Processes for Analyzing Positioned Trajectories in Sports\thanks{This work is an extension of the conference paper \citep{YFFFJ16}.}
}


\author{Yuxin~Zhao \and Feng~Yin \and Fredrik~Gunnarsson \and Fredrik~Hultkrantz 
        %etc.
}


\institute{Y.~Zhao and F.~Gunnarsson \at
              Ericsson AB, Datalinjen 4, Link\"{o}ping, SE-583 30, Sweden \\
%              Tel.: +46-725838257\\
%              Fax: +123-45-678910\\
              \email{firstname.lastname@ericsson.com}           %  \\
           \and
           F.~Yin \at 
           Department of Science and Engineering, Chinese University of Hong Kong, Shenzhen, CN-518172, China \\
           \email{yinfeng@cuhk.edu.cn}
           \and
           F.~Hultkrantz \at
           Ericsson AB, Mobilv\"{a}gen 12, Lund, SE-223 62, Sweden\\
           \email{fredrik.hultkrantz@ericsson.com}    
}

\date{Received: date / Accepted: date}

\maketitle

\begin{abstract}
	\label{sec:abs}
	Kernel-based machine learning approaches are gaining increasing interest for exploring and modeling large dataset in recent years. Gaussian process (GP) is one example of such kernel-based approaches, which can provide very good performance for nonlinear modeling problems. In this work, we first propose a grey-box modeling approach to analyze the forces in cross country skiing races. To be more precise, a disciplined set of kinetic motion model formulae is combined with data-driven Gaussian process regression model, which accounts for everything unknown in the system. Then, a modeling approach is proposed to analyze the kinetic flow of both individual and clusters of skiers. The proposed approaches can be generally applied to use cases where positioned trajectories and kinetic measurements are available. The proposed approaches are evaluated using data collected from the Falun Nordic World Ski Championships 2015, in particular the Men's cross country $4\times10$ km relay. Forces during the cross country skiing races are analyzed and compared. Velocity models for skiers at different competition stages are also evaluated. Finally, the comparisons between the grey-box and black-box approach are carried out, where the grey-box approach can reduce the predictive uncertainty by $30\%$ to $40\%$.   
\end{abstract}

\section{Introduction}
\label{sec:Introduction}
\subsection{Background}
Thanks to the emerging technologies in positioning, numerous positioned data are available and open to relevant research topics. Such data are usually analyzed to figure out different aspects that are hidden in the data, and machine learning is often used to explore the data distributions, study dependences between different attributes, provide modeling of patterns and make predictions. One specific use case is in sports, where the positions and kinetic measurements of athletes can be recorded. Such data analytics is crucial for both the coach and public audience, since they can provide valuable insights into the performance of athletes, which will further assist in deploying attacks and defenses in matches, monitoring physical condition of athletes and so on. %and , which include but not limited to Global Positioning System (GPS), inertial sensor assisted positioning, radio signal measurements based positioning, direct field sensing technology \citep{KSB17,FS16} and so on,

One interesting area is to use approaches in data analytics to analyze the kinetics of athletes in a certain sport, which may include studies on cause of motion, namely forces and torques, human movements with respect to the amount of time taken to carry out the activity and possible interactions between the athletes and their equipments and environments. All of these studies belong to the sports biomechanics research area, which are aiming to prevent injuries and improve performance of athletes. There have been numerous research conducted on sports biomechanics. For instance, \citep{BR07} provides a thorough introduction on sport biomechanics including the geometry of sports movements, forces in sport and how they are related by the law of kinetics and how these are related to the human body in a anatomical way. In \citep{PM14}, the effectiveness of different forces in cross country skiing are evaluated by conducting real measurements. However, most research focuses on explicit analysis from physical and biological points of view, and so far, the data-driven approaches, such as machine learning, haven't been fully explored. 

Another interesting topic in data analytics is the flow modeling and prediction. In literature, various studies arise around this area, which include human motion patterns \citep{ES09}, traffic flow modeling and prediction \citep{YZYD10}, moving patterns of a swarm of animals \citep{RMRR11}, and so on. Flow modeling and prediction can be done both on-line with real-time data, for instance the streaming probe data \citep{RJH10}, or off-line with cached/stored data, for instance the recorded data from surveillance cameras \citep{ES09}. However, to the best of our knowledge, these have never been explored under sport scenario, where a huge amount of positioned trajectories can be available.     

Historically, the most widely used machine learning approaches include neural networks \citep{smith94short}, support vector machines (SVM) \citep{zhang2008forecasting}, and Gaussian Processes (GP) \citep{MacKayGP97,RW06}. GP is one important class of kernel-based learning methods. First, it is good at exploring the relationship between a set of variables given a training dataset. Second, GP perfectly fits in the Bayesian framework, which allows for explicit probabilistic interpretation of model outputs. All these advantages make GP a powerful tool to address complex nonlinear regression and classification problems. However, the standard GP is computationally demanding when the dataset is large and its size grows with time. To remedy this drawback, a plethora of low-complex GP algorithms have been proposed over the last decade. Representative solutions include (1) reduced-rank approximations of the covariance matrix \citep{RW06} and (2) sparse representation of the complete training dataset \citep{CR05} and (3) partition of the complete dataset into smaller subsets and fusion of all local GP experts \citep{SNS06}, \citep{DN15} and (4) stochastic variational inference approximated GP \citep{Titsias09} and (5) recursive processing based GP including a grid based algorithm \citep{HMF13, Huber14} and a series of state-space model based algorithms \citep{SSH13}. In this paper, we narrow down our focus to the recursive processing based algorithms as they are more attractive for on-line applications. 

\subsection{Contribution}
In this work, we apply GP regression for modeling and prediction in a sport use case, but the proposed framework is generic for other use cases with positioned device trajectories and kinetic measurements, for instance, to model the temporal and/or spatial aspects of motions, speed, and data flows. In this work, we have used the data trajectories from the Falun Nordic World Ski Championships 2015 and the Men's cross country relay race, $4\times10$ km. The contribution of the work can be summarized as: 
\begin{enumerate}
	\item A grey-box modeling is proposed to perform force analysis. In a grey-box modeling approach, the internal working mechanism of the system is partially known. To be more specific, the force model in this work is formulated by combining the known deterministic motion kinetics with Gaussian processes regression to accounts for the unknown forces in skiing races. The model can be further used to investigate the performance of a specific skier and to study the differences between various skiing techniques. 
	\item A black-box modeling approach is proposed to model the ground speed. In the black-box modeling approach, the internal working mechanism of the system is completely unknown. In this work, both the standard and a grid based on-line Gaussian process regression \citep{HMF13} are proposed to provide a model specifying the relationships between the speed and position for each individual. For group of individuals, clustering is performed based on a number of features extracted from the training data. %Thereafter, a model for each cluster to estimate the conditional distribution over instantaneous speed given a specific position.  %As an on-line GP representative, we choose the gird based on-line GP  because it is adaptive to new-arrival data and, as will be shown later on the concept of updating prediction on fixed grids, fits the application specifically. at which an individual or a cluster of individuals are moving along predefined tracks, typically over multiple laps
	\item Not limited to ski race, the proposed approaches are also applicable to various sport activities, such as track and field, car race, horse race, and so on.  This can be utilized both on-line for private coaching/public use with real-time data and off-line for analysis with recorded data in batch.    
\end{enumerate} 
\subsection{Paper Organization and Notations}
The remainder of this paper is organized as follows: Section~\ref{sec:ForceAnalysis} proposes an approach for force analysis in sports, which combines the kinetics of motion with Gaussian processes. Section~\ref{sec:GP based flow modeling} introduces a modeling algorithm for positioned user trajectories based on both standard and the grid based on-line GP. Section~\ref{sec:ClusterFlow} introduces a novel strategy for clustering skiers based on selected features and the aggregated flow modeling for each cluster. Section~\ref{sec: DataDes} provides detailed descriptions of the dataset. Section~\ref{sec:Results} validates and compares the proposed algorithms in various scenarios with real data. Lastly, Section~\ref{sec:Conclusions} concludes the work. 
  
Throughout this paper, matrices are presented with uppercase letters and vectors with boldface lowercase letters. The operator $\left[\cdot\right]^{T}$ stands for vector/matrix transpose and $\left[\cdot\right]^{-1}$ stands for the inverse of a non-singular square matrix. The operator $\parallel \cdot \parallel$ stands for the Euclidean norm of a vector. $\mathcal{N}(\mu, \sigma^{2})$ denotes a Gaussian distribution with mean $\mu$ and variance $\sigma^{2}$. 

\section{Force Analysis for a Single Individual}
\label{sec:ForceAnalysis}
%For positioned data trajectories with kinetic measurements along the predefined tracks, it is possible to make use of the kinetic model of athletes. However, the motion of athlete as a complex process, there are always some factors that are unknown or uncertain.
The effective force applied by an athlete during a sports competition is complex to analyze. With global navigation satellite system (GNSS) trackers on the athlete, it is possible to obtain trajectories with time series of position and velocity estimates. Typically, the vertical position estimate from GNSS is uncertain. Instead, the horizontal position estimate can be used together with ground height information to estimate the vertical position. Based on kinetic relations, it is possible to model the athlete motion with the effective force as unknown. Hence, we propose to combine the information from kinetic models with Gaussian process regression to estimate the latent forces. To be more specific, we propose a generic way for analyzing forces of athletes at certain stages of the competition. 

\begin{figure*}[t]
	\centering
	\includegraphics[width=8.5cm]{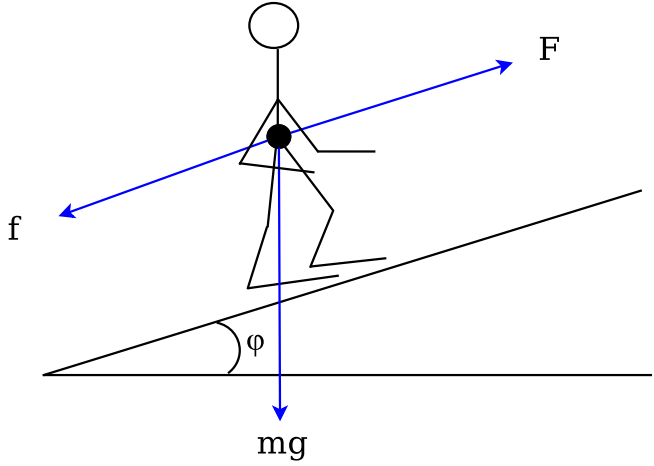}
	\caption{Force model for uphill.}
	\label{fig:fig1}
\end{figure*}

\begin{figure}[t]
	\centering
	\includegraphics[width=8.4cm]{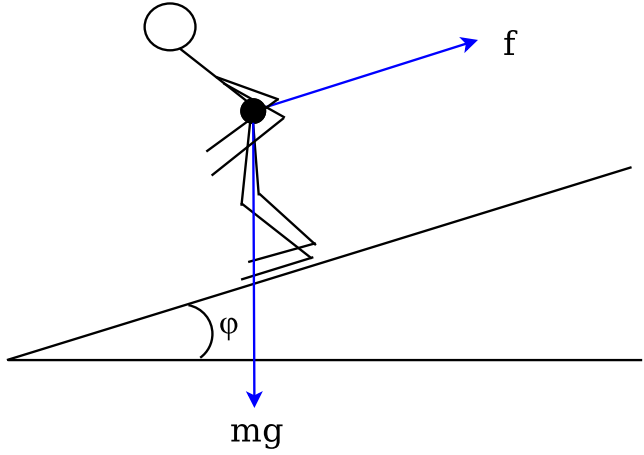}
	\caption{Force model for downhill.}
	\label{fig:fig2}
\end{figure}

We begin by analyzing a cross country skiing scenario, which can be easily extended to other sports with similar moving patterns. In cross country skiing, the skier moves forward by applying forces through poles and skis. In this process, there is also friction from the ice surface and the resistance from air. In order to ease the analysis, we propose the simplified force models as illustrated in Fig.~\ref{fig:fig1} and \ref{fig:fig2}. 

For uphill, there is a propulsive force $F$ going forward, while at the same time, $f$ in Fig.~\ref{fig:fig1} denotes both the friction from ice and the resistance from air, which is opposite to the moving direction. The mass of skier is denoted by $m$, and $g$ is the gravity of Earth and $\varphi$ is the track incline angle. In Fig.~\ref{fig:fig2}, the force model for downhill is illustrated, where usually there is no propulsive force, but only air resistance and friction from ice (which is denoted by $f$ in Fig.~\ref{fig:fig2}). The skier instead makes use of the gravity $mg$ to move forward with a decline angle of $\varphi$. The skier has to adjust his/her posture to reduce the air resistance. It should be noted that in a real practice, the forces are more complex than what have been illustrated in Fig.~\ref{fig:fig1} and \ref{fig:fig2}.

Based on the force models, in what follows, we analyze uphill and downhill scenarios, respectively. Since the forces are changing during different competition stages, the whole track is divided into small segments and we further assume the moving direction, the propulsive forces from a specific skier, air resistance and friction remain the same during each segment. In addition, the time for the skier to finish the segment is measured as $\Delta t$. Hence, according to Newton's law of motion, the change in velocity per time unit in an uphill segment can be formulate as 
\begin{equation}
\frac{\Delta \bm{v}}{\Delta t} m = F-\textrm{f}-mg\sin \varphi, \label{eq:motionLaw} 
\end{equation}
where the mass of skier is usually known and the incline angle $\varphi$ can be computed as the slope of the track is known. The change in velocity per time unit can be denoted as $a\triangleq\frac{\Delta \bm{v}}{\Delta t}$. In modern races, the velocity and position of athletes can be measured at certain fixed time stamps. However, it is typically difficult to measure the propulsive forces, since it comes from both skis and poles. Furthermore, the air resistance and friction on ice are almost impossible to measure, as they depend on many factors, such as the posture of skier, the temperature of ice, and the force from skier that is perpendicular to the track surface. Considering such complexities, we propose to model the resultant force, which is $F-f$, by a Gaussian process, which is formulated as
\begin{equation}
F_r(d) \triangleq F(d)-\textrm{f}(d) \triangleq r(d)+n_r,
\end{equation}  
where $d$ denotes the distance traveled from the beginning of the track, and it uniquely determines the position of skier when the track is predefined. $n_r$ is additive Gaussian noise with zero mean and variance $\sigma_{n_r}^2$. Hence, the full kinetic is given as 
\begin{equation}
\frac{\Delta \bm{v}}{\Delta t} m = -mg\sin \varphi +r(d)+n_r\label{eq:kineticGP}
\end{equation}
which consists of both the known gravity and the unknown forces. The function $r(d)$ follows a Gaussian process \cite{RW06}
\begin{equation}
r(d)\sim \mathcal{GP} (m_r(d), k_r(d,d')) \label{eq:GP_resultantForce}
\end{equation}
with mean $m_r(d)$ and kernel function $k_r(d,d')=\sigma_r^2\exp\left[-\frac{||d-d'||^2}{l_r^2}\right]$. In order to train the Gaussian process regression model, a set of training dataset is required, which is denoted by $\mathcal{D}\triangleq {(F_{r,1},d_1),\ldots,(F_{r,N},d_N)}$, where $F_{r,i}$ is constructed by $F_{r,i}\triangleq \frac{\Delta \bm{v}_i}{\Delta t_i}m+mg\sin \varphi_i$ for $i=1,\ldots,N$. The joint distribution of all observations $\bm{F}_r \triangleq [F_{r,1},\ldots, F_{r,N}]^T$ is given by 
\begin{equation}
p(\bm{F}_r|\bm{\theta}_r, \mathcal{D}) \sim \mathcal{N} (\mathbf{m}_r(\mathbf{d}), \mathbf{C}_r(\mathbf{d},\mathbf{d})), \label{eq:jointDistGPFA}
\end{equation}
where 
\begin{align}
\mathbf{d} & \triangleq [d_1, d_2, \ldots, d_N ]^T , \nonumber \\
\bm{\theta}_r &\triangleq [\sigma_r, l_r, \sigma_{n_r}]^T, \nonumber \\
\mathbf{m}_r(\mathbf{d})& \triangleq [m_r(d_1), m_r(d_2), \ldots, m_r(d_N)]^T , \nonumber \\
\mathbf{K}_r (\mathbf{d},\mathbf{d})  & \triangleq \begin{bmatrix}
k_r(d_1, d_1) & k_r(d_1, d_2)  & \ldots & k_r(d_1, d_{N})         \\[0.2em]
%k(d_2, d_1) & k(d_2, d_2)  & \ldots & k(d_2, d_{M})        \\[0.2em]
\vdots & \vdots & \ddots & \vdots         \\[0.3em]
k_r(d_{N}, d_1) & k_r(d_{N}, d_2) & \ldots & k_r(d_{N}, d_{N})         \\[0.2em]
\end{bmatrix}, \nonumber \\
\mathbf{C}_r(\mathbf{d}, \mathbf{d}) & \triangleq \mathbf{K}_r(\mathbf{d}, \mathbf{d}) + \sigma_{n_r}^2 \mathbf{I}_{N}.  \nonumber
\end{align}
The parameters $\bm{\theta}_r$ can be estimated by maximizing the likelihood function given in \eqref{eq:jointDistGPFA}. The detailed explanations are given in Appendix A. Given a new input location $d_*$ and the training set $\mathcal{D}$, the resultant force can be estimated by 
\begin{equation}
F_r(d_*)|\mathcal{D} \sim \mathcal{N}(\hat{\mu}_r(d_*),\hat{\sigma}_r(d_*) ),
\end{equation}
where 
\begin{subequations}
	\begin{align}
	\hat{\mu}_r(d_*)&= \mathbf{k}_r^{T}(d_{*}, \mathbf{d}) \mathbf{C}_r^{-1}(\mathbf{d}, \mathbf{d}) (\bm{F}_r - \mathbf{m}_r(\mathbf{d})) + m_r(d_{*}),\\
	\hat{\sigma}_r^2(d_*)&= \sigma_{n_r}^2 + \sigma_{r}^2 - \mathbf{k}_r^{T}(d_{*}, \mathbf{d}) \mathbf{C}_r^{-1}(\mathbf{d}, \mathbf{d}) \mathbf{k}_r(d_{*}, \mathbf{d}).
	\end{align}
\end{subequations}

To analyze the downhill segment, as shown in Fig.~\ref{fig:fig2}, the following holds according to the law of motion:
\begin{align}
\frac{\Delta \bm{v}}{\Delta t} m = mg\sin \varphi-\textrm{f}, \label{eq:motionLawDH} 
\end{align}
where $F_r\triangleq-\textrm{f}$ is modeled by 
\begin{equation}
F_r(d) = r(d)+n_r,
\end{equation}
and $r(d)$ follows a Gaussian process as given in \eqref{eq:GP_resultantForce}. Then, we follow the same procedure applied to estimate the resultant force for uphill. Instead, the training observations $F_{r,i}$ for $i=1,\ldots, N$ are constructed as $F_{r,i}=\frac{\Delta \bm{v}_i}{\Delta t_i} m-mg\sin \varphi_i$. 

\section{GP Based Flow Modeling and Prediction for a Single Individual}
\label{sec:GP based flow modeling}

In previous section, we have proposed a grey-box modeling approach for analyzing the forces in, for instance, skiing races. The grey-box modeling approach explores the known kinetic model based on the physical laws and combines it with Gaussian process regression models which are used to account for the unknown forces. It is also possible to perform the modeling of input and output relationship by a black-box approach, where it is not necessary to know explicitly the internal working mechanics. In this section, we propose a black-box approach to analyze the relationship between the ground speed and the position of an individual skier. %In this section, we will first review the standard Gaussian process which will be applied to our flow model. Before introducing the GP method, the flow model we wish to learn is provided.
%\subsection{Flow Model}
%\label{subsec:flowModel}
%
%Denoting by $\mathbf{x}_t$ the 2-dimensional position of an individual at a sampling time instance $t$, the change in position from time $t$ to $t+1$ can be expressed as
%%
%\begin{equation}
%\mathbf{x}_{t+1}-\mathbf{x}_t = g(\mathbf{x}_t)+ \mathbf {n}, 
%\label{eq: motion}
%\end{equation}
%%
%where $g(\mathbf{x}_t)$ is a flow model and $\mathbf {n}$ is additive noise. If we denote the change in position as $\Delta \mathbf{x}_t = \mathbf{x}_{t+1}-\mathbf{x}_{t}$ and divide both sides of (\ref{eq: motion}) by the time difference $\Delta t$, we will have the velocity $\mathbf{v}_t \triangleq \frac{\Delta \mathbf{x}_t}{\Delta t}$ as a function of $\mathbf{x}_t$. Instead of working on a multiple output spatio-temporal GP of the velocity, we consider a simpler, spatial GP with scalar output.  
%
%\begin{equation}
%\mathbf{v}_t = g(\mathbf{x}_{t})+\mathbf{e},  \nonumber
%\end{equation}
%
%where $\mathbf{v}_t = \frac{\Delta \mathbf{x}_t}{\Delta t}$, $g(\mathbf{x}_{t}) = \frac{g^\prime(\mathbf{x}_{t})}{\Delta t}$ and $\mathbf{e}$ is some additive noise. 

Concretely, we estimate the ground speed,  $v_t = \parallel \mathbf{v}_t \parallel$, for a single skier at a specific position. Since the individual follows a predefined track in this scenario, the position of an individual at time $t$ can be uniquely translated into the distance traveled on the track since the start of the race, denoted by $d_t$ herein. With the definitions given above, the following flow model is formulated
\begin{equation}
v_t(d) = f(d) + n, 
\label{eq: speedmodel}
\end{equation}  
where $f(\cdotp)$ is the underlying flow model and $n$ is additive noise, which is assumed to be Gaussian distributed with zero mean and variance $\sigma_n^2$. The focus of this section is to use GP regression to infer the underlying flow model in (\ref{eq: speedmodel}) and predict the ground speed value at any input $d_*$.  
\subsection{Standard Gaussian Process Regression}
\label{subsec: full GP}
In this subsection, the standard Gaussian process (SGP) will be introduced and applied to the problem formulated above. Without specifying the time, the previously defined function $f$ can be approximated by a GP, which is given by 
\begin{equation}
f(d) \sim \mathcal{GP}(m(d), k(d,d^\prime)), \nonumber
\end{equation}
where $m(d) = v_0$
%\begin{equation}
%m(d) = v_0 \nonumber
%\end{equation} 
is the mean function (we assume $v_0 = 0$ in this work) and $k(d,d^\prime)$ is the covariance/kernel function. \\

In the training phase, a dataset denoted as, $\mathcal{S} = \left\lbrace (d_1, v_1), \ldots, (d_M, v_M) \right\rbrace$, is collected. Considering the additive noise, the joint distribution of the observed ground speed measured at different distances is given by %\footnote{With a bit abuse of notation, $\mathbf{v}$ is a vector of ground speed values measured for different distances on track.}
\begin{equation}
p(\mathbf{v} (\mathbf{d})|\mathcal{S}) \sim \mathcal{N} (\mathbf{m} (\mathbf{d}), \mathbf{C} (\mathbf{d}, \mathbf{d})), \nonumber
\label{GPdistribution}
\end{equation} 
%
%with the following notations introduced: 
%%
%\begin{align}
%%
%\mathbf{d} & \triangleq [d_1, d_2, \ldots, d_M ]^T , \nonumber \\
%%
%\mathbf{v}(\mathbf{d}) & \triangleq [v_1 , v_2 , \ldots, v_M ]^T  , \nonumber \\
%%
%\mathbf{m}(\mathbf{d}) & \triangleq [m(d_1), m(d_2), \ldots, m(d_M)]^T , \nonumber \\
%%
%\mathbf{K}(\mathbf{d}, \mathbf{d}) & \triangleq \begin{bmatrix}
%k(d_1, d_1) & k(d_1, d_2)  & \ldots & k(d_1, d_{M})         \\[0.2em]
%%k(d_2, d_1) & k(d_2, d_2)  & \ldots & k(d_2, d_{M})        \\[0.2em]
%\vdots & \vdots & \ddots & \vdots         \\[0.3em]
%k(d_{M}, d_1) & k(d_{M}, d_2) & \ldots & k(d_{M}, d_{M})         \\[0.2em]
%\end{bmatrix}, \nonumber \\
%%
%\mathbf{C}(\mathbf{d}, \mathbf{d}) & \triangleq \mathbf{K}(\mathbf{d}, \mathbf{d}) + \sigma_{n}^2 \mathbf{I}_{M}.  \nonumber
%%
%\end{align}    
%
where $\mathbf{d}  \triangleq [d_1, d_2, \ldots, d_M ]^T $, and $\mathbf{v} (\mathbf{d})$, $\mathbf{m} (\mathbf{d})$ and $\mathbf{C} (\mathbf{d}, \mathbf{d})$ can be easily constructed as given in \eqref{eq:jointDistGPFA}. When there comes a novel value of $d_*$, %the joint distribution of the observed target values and the predict value $v_*$ at $d_*$ is given by
%
%\begin{equation}
%\begin{bmatrix}
%\mathbf{v} \\[0.2em]
%v_* \\[0.2em]
%\end{bmatrix} \sim \mathcal{N} \left(\mathbf{0}, \begin{bmatrix}
%\mathbf{C} (\mathbf{d}, \mathbf{d}) & \mathbf{k}(d_*, \mathbf{d}) \\
%\mathbf{k}^{T}(d_*, \mathbf{d}) & k(d_*, d_*) \\
%\end{bmatrix}   \right),
%\label{eq: pre_distribution}
%\end{equation}
%%
%where $k(d_*, \mathbf{d}) \triangleq [k(d_*, d_1), k(d_*, d_2), \ldots, k(d_*, d_M)]^T$. Then, 
we compute according to \citep{RW06} the Gaussian posterior probability of a ground speed value at a new $d_*$ by
\begin{equation}
p(v(d_{*})| \mathcal{S}) \sim \mathcal{N}\left(\hat{\mu}(d_{*}), \hat{\sigma}^{2}(d_{*}) \right),
\label{eq:GP-posterior}
\end{equation} 
where
\begin{subequations}
	\begin{equation}
	\hat{\mu}(d_{*}) = \mathbf{k}^{T}(d_{*}, \mathbf{d}) \mathbf{C}^{-1}(\mathbf{d}, \mathbf{d}) (\mathbf{v}(\mathbf{d}) - \mathbf{m}(\mathbf{d})) + m(d_{*}),
	\label{eq:predictedMean}
	\end{equation}
	\begin{equation}
	\hat{\sigma}^{2}(d_{*}) = \sigma_{n}^2 + k(d_*, d_*) - \mathbf{k}^{T}(d_{*}, \mathbf{d}) \mathbf{C}^{-1}(\mathbf{d}, \mathbf{d}) \mathbf{k}(d_{*}, \mathbf{d}),
	\label{eq:predictedCov}
	\end{equation}
\end{subequations}
and $ k(d_*, \mathbf{d}) \triangleq [k(d_*, d_1), k(d_*, d_2), \ldots, k(d_*, d_M)]^T$.

The SGP deals with the training data in a batch manner. The corresponding computational complexity scales as $\mathcal{O}(M^3)$ and the memory requirement scales as $\mathcal{O}(M^2)$. Next, we apply the grid based on-line Gaussian process (OGP) \citep{HMF13} to derive an on-line ground speed model.
%
%Recently, some sparse GPR methods have been proposed to reduce the computational complexity to $\mathcal{O}(s^{2}M)$ and the memory requirement to $\mathcal{O}(sM)$, where $s$ ($s \ll M$) is the size of an additional set of inducing input points that can be selected randomly from the training dataset or according to some information criteria. A unifying view of the state-of-the-art sparse GPR methods has been given in \citep{CR05}. 
%
\subsection{Grid Based On-line Gaussian Process Regression}
\label{subsec: OL_GP}
The notations, if not re-defined, will follow those used for the SGP. For simplicity and easier comparison with the SGP, we imagine that the training data arrives one by one in time, namely we have a new data point $\{ d_t, v(d_{t}) \}$, at time instance $t=1,2,\ldots,M$.  \\

In grid based on-line GP, a set of grids $\bar{\mathbf{d}} = \left[ \bar{d}_1, \bar{d}_2, \ldots, \bar{d}_s \right]$ is introduced to represent some predefined reference distances on track. The corresponding ``clean'' ground speed (without the additive white Gaussian noise) values at these grids are latent variables $\bar{\mathbf{v}}(\bar{\mathbf{d}}) \triangleq \left[ \bar{v}(\bar{d}_1), \bar{v}(\bar{d}_2), \ldots, \bar{v}(\bar{d}_s) \right]^T$. We denote $\mathcal{S}_g \triangleq \{\bar{\mathbf{v}}(\bar{\mathbf{d}}), \bar{\mathbf{d}}\}$. For notational brevity in the sequel, $\bar{\mathbf{v}}$ is short for $\bar{\mathbf{v}}(\bar{\mathbf{d}})$, and its mean and covariance matrix are denoted by $\bar{\mathbf{m}}$ and $\bar{\mathbf{K}}$, respectively. Our aim is to compute the posterior distribution of $\bar{\mathbf{v}}$ at any time instance $t$ ($t \geq 1$) given the training data $\mathcal{S}_{1:t} \triangleq \left\lbrace (d_1, v_1), \ldots, (d_t, v_t) \right\rbrace$. The main steps of the grid based OGP \citep{HMF13} are summarized as follows:
%Herein, the noisy ground speed observations are stacked into $\mathbf{v}(\mathbf{d}_{1:t}) \triangleq [v(d_{1}), v(d_{2}), \ldots, v(d_{t})]^T$, and the distances are stacked into $\mathbf{d}_{1:t} \triangleq [d_{1}, d_{2}, \ldots, d_{t}]$.
%
\begin{enumerate}
	\item \textit{Initialization}: Set initial mean vector $\bm{\mu}_{0}^{g} \triangleq \bar{\mathbf{m}}$ and the covariance matrix $\mathbf{K}_{0}^{g} \triangleq \bar{\mathbf{K}}$. Compute the inverse of $\bar{\mathbf{K}}$ and store it for use later on. Here the prior mean $\bar{\mathbf{m}}$ is set to be a vector of all zeros (of size $s$) and the prior covariance matrix is set to be
	%\begin{equation}
	%\bar{\mathbf{m}} = [0, 0, \ldots, 0]^T, \nonumber
	%\end{equation}
	%
	\begin{equation}
	\bar{\mathbf{K}} = \begin{bmatrix}
	k(\bar{d}_1, \bar{d}_1) & k(\bar{d}_1, \bar{d}_2)  & \ldots & k(\bar{d}_1, \bar{d}_{s})         \\[0.2em]
	%k(\bar{d}_2, \bar{d}_1) & k(\bar{d}_2, \bar{d}_2)  & \ldots & k(\bar{d}_2, \bar{d}_{s})        \\[0.2em]
	\vdots & \vdots & \ddots & \vdots         \\[0.3em]
	k(\bar{d}_{s}, \bar{d}_1) & k(\bar{d}_{s}, \bar{d}_2) & \ldots & k(\bar{d}_{s}, \bar{d}_{s})         \\[0.2em]
	\end{bmatrix}.
	\end{equation}
	\item \textit{Recursive Processing}: For each $t = 1,2,\ldots,M$, do the following computations:
	\begin{subequations}
		\begin{align}
		\mathbf{J}_{t}  &=  \mathbf{k}(d_{t}, \bar{\mathbf{d}}) \bar{\mathbf{K}}^{-1} \label{eq:onlineGPrecStart}, \\
		\mu_{t}^{p}  &=  m(d_{t}) + \mathbf{J}_{t} \!\left( \bm{\mu}_{t-1}^{g} - \bar{\mathbf{m}} \right), \\
		\sigma_t^{2,p}  &= k(d_t, d_t)  + \mathbf{J}_{t} \!\left( \mathbf{K}_{t-1}^{g} - \bar{\mathbf{K}} \right)\! \mathbf{J}_{t}^{T}, \\
		\tilde{\mathbf{g}}_{t}  &=  \frac{1}{\sigma_{n}^2 + \sigma_t^{2,p}} \mathbf{K}_{t-1}^{g} \mathbf{J}_{t}^{T}, \\
		\bm{\mu}_{t}^{g}  &=  \bm{\mu}_{t-1}^{g} + \tilde{\mathbf{g}}_{t} \!\left( v(d_{t}) - \mu_{t}^{p} \right), \\
		\mathbf{K}_{t}^{g}  &=  \mathbf{K}_{t-1}^{g} - \tilde{\mathbf{g}}_{t} \mathbf{J}_{t} \mathbf{K}_{t-1}^{g}. \label{eq:onlineGPrecEnd} 
		\end{align} 
	\end{subequations}
	After the recursive processing through (\ref{eq:onlineGPrecStart})--(\ref{eq:onlineGPrecEnd}), we have 
	\begin{equation}
	p(\bar{\mathbf{v}} | \bar{\mathbf{d}}, \mathcal{S}) = \mathcal{N} \left( \bar{\mathbf{v}} | \bm{\mu}_{M}^{g}, \mathbf{K}_{M}^{g} \right).
	\end{equation}
	\item \textit{Prediction}: At the end of the training phase, namely $t=M$ assumed in this specific example, the posterior distribution of a noisy speed observation $v({d}_{*})$ at a novel input position $d_{*}$, given $\mathcal{S}$ and $\mathcal{S}_g$, can be approximated by 
	\begin{equation}
	p( v(d_{*})| \bar{\mathbf{d}}, \mathcal{S} ) \approx \mathcal{N}(v(d_{*})| \hat{\mu}(d_{*}), \hat{\sigma}^{2}(d_{*}) ),  
	\label{eq:onlineGP-posterior}
	\end{equation}
	where
	\begin{subequations}
		\begin{equation}
		\hat{\mu}(d_{*}) \!=\! \bar{\mathbf{k}}^{T}(d_{*}) \bar{\mathbf{K}}^{-1} (\bm{\mu}_{M}^{g} - \bar{\mathbf{m}}) + m(d_{*}), 
		\label{eq:mu-onlineGP}
		\end{equation}
		\begin{equation}
		\hat{\sigma}^{2}(d_{*}) \!=\! k(d_{*}) + \sigma_{n}^{2} + \bar{\mathbf{k}}^{T}\!(d_{*}) \bar{\mathbf{K}}^{-1} \!\! \left(\! \mathbf{K}_{M}^{g} \bar{\mathbf{K}}^{-1} \!-\! \mathbf{I}_s \!\right)\! \bar{\mathbf{k}}(d_{*}).
		\label{eq:var-onlineGP}
		\end{equation}
	\end{subequations}
	Herein, $\bar{\mathbf{k}}(d_{*})$ is short for $\mathbf{k}(d_{*}, \bar{\mathbf{d}})$ and $k(d_{*})$ is short for $k(d_{*}, d_{*})$.
\end{enumerate}	
 The detailed derivations of (\ref{eq:onlineGPrecStart})--(\ref{eq:onlineGPrecEnd}) can be found in \citep{HMF13} and the derivations of (\ref{eq:mu-onlineGP}) and (\ref{eq:var-onlineGP}) are given in Appendix B. It is easy to verify that the computational complexity scales as $\mathcal{O}(s^3)$ for $\bar{\mathbf{K}}^{-1}$ in the initialization step, $\mathcal{O}(s^2)$ for $\bm{\mu}_{t}^{g}$ and $\mathbf{K}_{t}^{g}$ at any time instance $t$ in the recursive processing step. The computational complexity of $\bm{\mu}_{M}^{g}$ and $\mathbf{K}_{M}^{g}$ scales as $\mathcal{O}(s^2 M)$. The computational complexity for prediction in the third step scales as $\mathcal{O}(s^2)$. As compared to the SGP, the grid based OGP is able to reduce the overall computational complexity from $\mathcal{O}(M^3)$ to $\mathcal{O}(s^2 M)$ with $s \ll M$. Moreover, when we have a new observation pair $\{ d_{M+1}, v(d_{M+1}) \}$ at time $M+1$ after the training phase, it requires only $\mathcal{O}(s^2)$ complexity to compute $\bm{\mu}_{M+1}^{g}$ and $\mathbf{K}_{M+1}^{g}$, which is essential for on-line learning.     

Apart form the reduced computational complexity, there are several other benefits of using OGP as compared to the SGP. For instance, model fitting can be performed in parallel to measurement collection, and we can stop collecting more data when the posterior distribution of ground speed at the predefined grids converges. To summarize, OGP is more flexible to use and more adaptive to new arrival data. While if the underlying model is time invariant and the computational cost is secondary, the SGP using all available training data for both hyper-parameter optimization and prediction will intuitively give the best modeling results.        
\subsection{Kernel Selection}
\label{subsec: kernels}
Kernel function is a key component of GP, as it encodes the assumptions about the function which we wish to learn. The kernel function reflects the similarity between data points \citep{RW06}. In this subsection, the selection of different kernels will be discussed. One classic kernel function is the Squared Exponential (SE) kernel, defined by
\begin{equation}
k(d, d^\prime) = \sigma_s^2 \exp \left[ -\frac{(d-d^\prime)^2}{l_d^2} \right], \nonumber
\end{equation}
where $\sigma_s^2$ is the variance of the function and $l_d$ is the length scale which determines how rapidly the function varies with $d$. The SE kernel is considered as the most widely used kernel. However, it implies a stationary model which forbids structured extrapolation \citep{KZSH16}. In some specific cases, this kernel function may show poor performance in prediction, for instance, in sport races where there is periodic pattern over laps. Considering this, it is more appropriate to adopt a periodic kernel which can reflect the similarities between different laps. However, strict periodicity is too rigid, because there may be some deviations in each lap (e.g., due to the strength loss of the individual, strategies used in competition, etc). Hence, we adopt a local periodic (LP) kernel, which is a product of an SE kernel and a periodic kernel: 
\begin{equation}
k(d, d^\prime) = \sigma_s^2 \exp \left[ -\frac{\sin^2\left(\frac{\pi (d-d^\prime) }{\lambda}\right)}{l_p^2} \right] \exp \left[ -\frac{{( d-d^\prime)}^2}{l_d^2} \right],
\label{eq:localPeriodicKernel}
\end{equation} 
where $l_p$ is the length scale of the periodic kernel and $\lambda$ is the period-length. This kernel considers two inputs are similar if they are similar under both the SE and the periodic kernels. If $l_d \gg \lambda$, this allows encoding a decay in the covariance over several oscillations \citep{KZSH16}. The key benefits of this kernel is that it outperforms SE kernel in prediction when the distance from the data is increasing as illustrated in \citep[Fig. 4]{KZSH16}.  
%Since, the product of two kernel functions is also a kernel and therefore a valid covariance function for a GP \citep{RW06}. 
%

 Lastly, we note that the LP kernel in (\ref{eq:localPeriodicKernel}) is not necessary optimal in our application, but as will be shown in our simulations it gives very good modeling and prediction results. Interested readers can refer \citep{DLG13, WA13,Yin18} for strategies for selecting an optimal kernel from the training data.   
\subsection{Hyperparameters Determination}
\label{subsec: Hyperparas}
Given the SGP model and the kernel in \ref{subsec: full GP} and \ref{subsec: kernels} respectively, the hyperparameters to be calibrated are
\begin{equation}
\bm{\theta} \triangleq [\sigma_n^2, \sigma_s^2, l_p, l_d]^T. \nonumber
\end{equation}
The likelihood function of the observed ground speed with respect to the hyperparameters $\bm{\theta}$ can be written as follows:
\begin{equation}
p(\mathbf{v}(\mathbf{d}); \bm{\theta}) \sim \mathcal{N}({m}(\mathbf{d}), \mathbf{C}(\mathbf{d}, \mathbf{d}; \bm{\theta})).  
\label{eq:LikelihoodGP}
\end{equation} 
Here, the maximum-likelihood estimate (MLE), $\hat{\bm{\theta}}$, is derived. Details are given in Appendix A.

 In the OGP, we assumed that the parameters $\bm{\theta}$ are known before the recursive process starts. This can be the case when some historical/expert knowledge is available or a small set of the training data can be used to train the parameters like we did for the SGP. Huber demonstrated in \citep{Huber14} that these parameters can be learned on-line as well.

\section{Aggregated Flow Modeling And Prediction for Multiple Individuals}
\label{sec:ClusterFlow}
In this section, we investigate aggregated flow modeling and prediction for multiple individuals that are clustered. 
%
%\subsection{A Brief Overview of Sequence Clustering}
%\label{sub:seqCluster}
%
The classic way of clustering data sequences is to extract some common features from the data sequences and then perform K-means algorithm or expectation-maximization (EM) algorithm \citep{Bishop06} based upon some distance metric. Principle component analysis \citep{Bishop06} can be used to reduce the dimension of the feature space before running the K-means or EM algorithm. The drawback of these classic methods is that the number of clusters need to be prescribed before we conduct clustering. A more sophisticated way is to combine the Dirichlet process with Gaussian processes in a Bayesian framework, which is capable of modeling an infinite number mixture stochastic processes (see for instance \citep{HRL15}). One example of sequence clustering will be given in Section \ref{subsec: AggResults}.           
\subsection{Flow Modeling and Prediction for Multiple Individuals}
\label{subsec: FlowMulti}
%In previous subsection, the clustering of individuals is applied based on various features extracted from the training data. 
%This section introduces the aggregated flow modeling and prediction for a cluster of individuals. Clustering of data sequences can be performed both in spatial/temporal domain (see for instance \citep{WMG12}) or in frequency domain (see for instance \citep{Lin2004}). 
The data of all individuals in the same cluster will be aggregated to form a new dataset, denoted as $D = \left\lbrace (v_1, d_1), (v_2, d_2), \ldots, (v_{N_D}, d_{N_D}) \right\rbrace$. The ground speed will be modeled as a function of the distance on track plus some noise terms:
\begin{equation}
v(d) = h(d)+n_c+n_w
\label{eq:cFlowModel}
\end{equation}
where $n_w$ is an additive white Gaussian noise with zero mean and variance $\sigma_n^2$, and $n_c$ is an additive correlated Gaussian noise with zero mean and variance $\sigma_c^2$. However, $n_c$ at two positions, namely $d$ and $d^\prime$, is assumed to be correlate, accounting for the interactive effects between individuals. The kernel function is selected as $k_c(d,d')\triangleq \sigma_{c}^{2} \exp\left[ \frac{-( d - d' )^2}{l_{c}^2} \right]$. Compared to the flow model for an individual, the correlations between individuals in one cluster are important since their performance may affect each other during the competition. Such effects can be considered as correlated noise and thus be modeled as given in \eqref{eq:cFlowModel}.
Let $h(d)$ be a GP with the mean function $m(d)$ and the kernel function $k(d,d')=\sigma_s^2 \exp (-\frac{{( d-d^\prime )}^2}{l_d^2})$, the joint distribution of observations is given by
\begin{equation}
p(\mathbf{v} (\mathbf{d})|\mathcal{D}; \bm{\theta}_c) \sim \mathcal{N} (\mathbf{m}_c (\mathbf{d}), \mathbf{C}_c (\mathbf{d}, \mathbf{d})), \nonumber
\label{eq:GPdistributionCluster}
\end{equation}
where $\bm{\theta}_c  = [\sigma_n^2, \sigma_s^2, l_d, \sigma_c^2, l_c]^T$ and
%$\bm{\theta}_c = [\sigma_n^2, \sigma_s^2, l_d, \sigma_c^2, l_c]^T$, $\mathbf{d} \triangleq [d_1, d_2, \ldots, d_{N_D} ]^T$, and $\mathbf{v}(\mathbf{d})$, $\mathbf{m}_c(\mathbf{d})$ and $\mathbf{C}_c(\mathbf{d}, \mathbf{d})$ can be constructed as in \eqref{eq:jointDistGPFA}
\begin{align}
%\bm{\theta}_c & = [\sigma_n^2, \sigma_s^2, l_d, \sigma_c^2, l_c]^T, \nonumber \\
%
\mathbf{d} & \triangleq [d_1, d_2, \ldots, d_{N_D} ]^T , \nonumber \\
\mathbf{v}(\mathbf{d}) & \triangleq [v_1 , v_2 , \ldots, v_{N_D} ]^T  , \nonumber \\
\mathbf{m}_c(\mathbf{d}) & \triangleq [m(d_1), m(d_2), \ldots, m(d_{N_D})]^T , \nonumber \\
\mathbf{K}_c(\mathbf{d}, \mathbf{d}) & \triangleq \begin{bmatrix}
k'(d_1, d_1) & k'(d_1, d_2) & \ldots & k'(d_1, d_{{N_D}})         \\[0.2em]
%k'(d_2, d_1) & k'(d_2, d_2)  & \ldots & k'(d_2, d_{{N_D}})        \\[0.2em]
\vdots & \vdots & \ddots & \vdots         \\[0.3em]
k'(d_{N_D}, d_1) & k'(d_{N_D}, d_2) & \ldots & k'(d_{N_D}, d_{N_D})        \\[0.2em]
\end{bmatrix}, \nonumber \\
\mathbf{C}_c(\mathbf{d}, \mathbf{d}) & \triangleq \mathbf{K}_c(\mathbf{d}, \mathbf{d}) + \sigma_{n}^2 \mathbf{I}_{N_D},  \nonumber
\end{align}
and $k'(d, d') = k(d,d')+k_c(d,d')$. Correspondingly, the posterior probability of an observed ground speed value $v_*$ at a novel input $d_*$ is given by
\begin{equation}
p(v(d_{*})| \mathcal{D}) \sim \mathcal{N}\left(\hat{\mu}_c(d_{*}), \hat{\sigma}_c(d_{*}) \right),
\label{eq:GPposteriorCluster}
\end{equation} 
where
\begin{subequations}
	\begin{equation}
	\hat{\mu}_c(d_{*}) = \mathbf{k}'^{T}(d_{*},\mathbf{d})\mathbf{C}_c^{-1}(\mathbf{d}, \mathbf{d}) (\mathbf{v}(\mathbf{d}) - \mathbf{m}_c(\mathbf{d})) + m(d_{*}),
	\label{eq:predictedMeanCluster}
	\end{equation}
	\begin{equation}
	\hat{\sigma}_c^2(d_{*}) = \sigma_{n}^2 + \sigma_{c}^2 + \sigma_{s}^2 - \mathbf{k}'^{T}(d_{*},\mathbf{d}) \mathbf{C}_c^{-1}(\mathbf{d},\mathbf{d})\mathbf{k}'^{T}(d_{*},\mathbf{d}), 
	\label{eq:predictedCovCluster}
	\end{equation}
\end{subequations}
%
%and $k(d_*)=k(d_*,d_*)$, $k_c(d_*)=k_c(d_*, d_*)$,
%
%\begin{align}
%\mathbf{k}'^{T}(d_{*},\mathbf{d})&=\mathbf{k}^{T}(d_{*},\mathbf{d})+\mathbf{k}_c^{T}(d_{*},\mathbf{d}) \nonumber \\
% &= [k(d_*, d_1), k_c(d_*, d_2), \ldots, k_c(d_*, d_{N_D})]^T \nonumber \\
% &+[k_c(d_*, d_1), k_c(d_*, d_2), \ldots, k_c(d_*, d_{N_D})]^T. \nonumber
%\end{align} 
%
Similarly, the MLE method is applied to train the hyperparameters $\hat{\bm{\theta}}_c$, and more details are given in Appendix A. 
\begin{figure}[t]
	\centering
	\includegraphics[width=8.5cm]{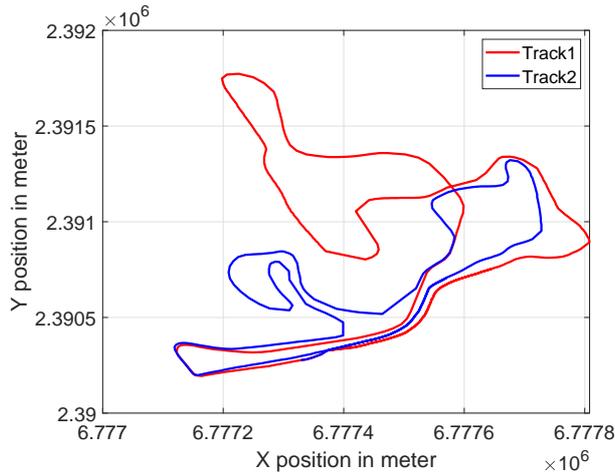}
	\caption{Map of two tracks.}
	\label{fig:figure3}
\end{figure} 
\begin{figure}[t]
	\centering
	\subfigure[]{
		\includegraphics[width=8.2cm,height=4.2cm]{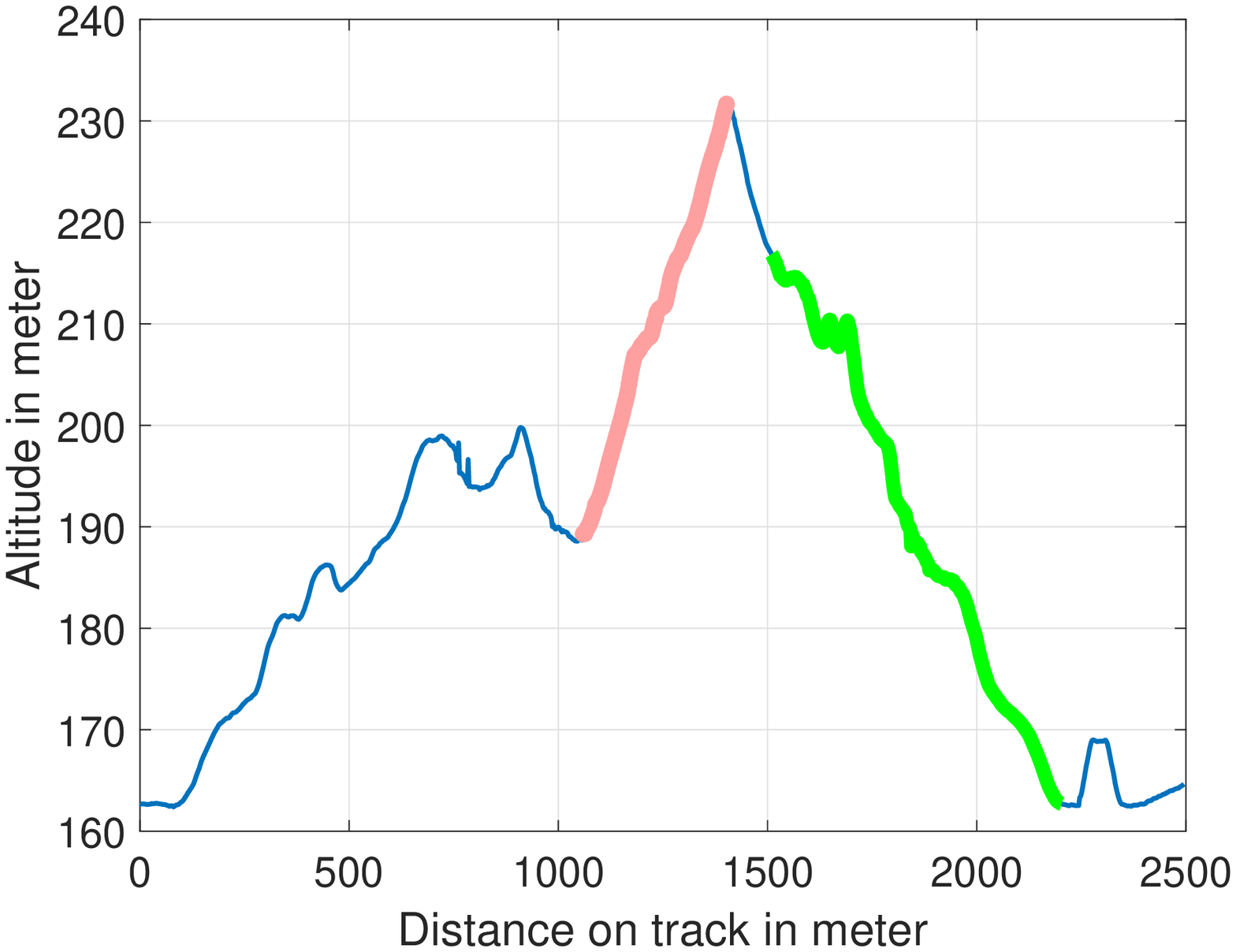}
		\label{fig:figure1}}
	\subfigure[]{
		\includegraphics[width=8.2cm,height=4.2cm]{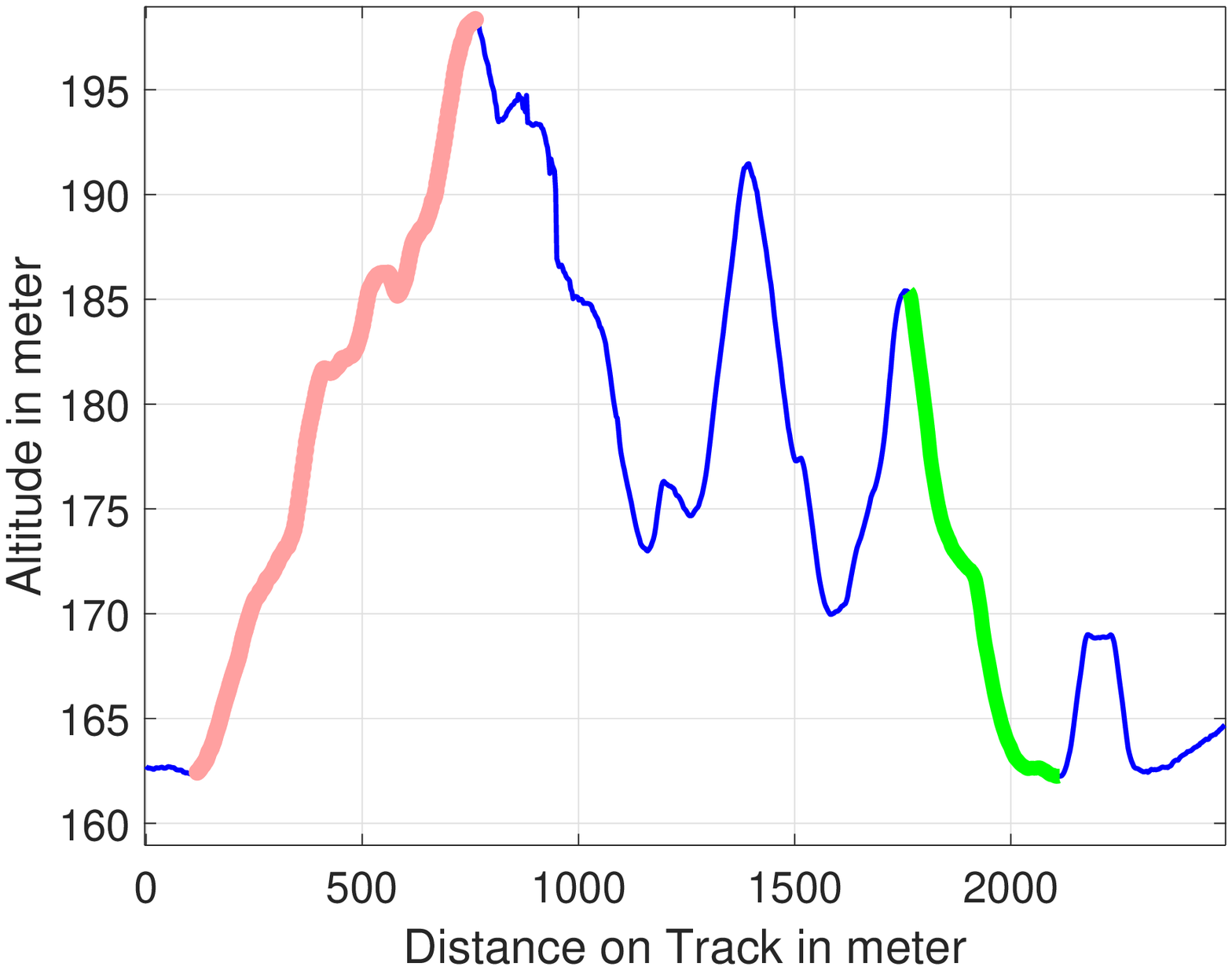}
		\label{fig:figure1a}}
	\caption{Altitude maps for track 1 and 2.}
\end{figure}

\section{Data Description}
\label{sec: DataDes}
%
%\subsection{Data Description}
%\label{subsec:Description}
%
The data used in this work was gathered during the men's $4 \times 10$ kilometers relay race in the 2015 Nordic World Ski Championships, Falun, Sweden. The dataset contains primarily the longitude, latitude, distance on track, ground speed and sampling time instances for $56$ individuals from $17$ national teams. The individuals who did not finish the competition have been excluded from the dataset. The data was conducted regularly at a frequency of 1 Hz. A distance on track is essentially calculated by \textregistered TrackingMaster from the longitude and latitude of a skier obtained from global positioning system (GPS). \textregistered TrackingMaster is a software developed by Swiss Timing to receive raw GPS data from GPS modules and to convert the raw GPS data into data related to the course. The positioning uncertainty of GPS in a wide open outdoor environment can be as low as 5 meters, hence is ignored in our work. The individuals compete on track $1$ for relay $1$ and $2$, while on track $2$ for relay $3$ and $4$. The two tracks are illustrated in Fig. \ref{fig:figure3} with their coordinates expressed using the World Geodetic System (WGS). Each track is 2.5 kilometers in length. In each relay, an individual has to finish $10$ kilometers on one track (i.e., $4$ laps). The length of data is different due to the various finishing time between individuals. It should be noted that on track 1, skiers are applying the \textit{classic style}, while on track 2 they use \textit{skating style}. 

 Altitude maps, see for instance Fig.\ref{fig:figure1} and \ref{fig:figure1a}, are readily available before the race. For individual force analysis, we apply the approach to the \textit{killer hill} (i.e., highlighted with red color) and \textit{steepest downhill} segments (i.e., highlighted with green color). For flow model of multiple individuals, the data of each individual is first segmented. Then, the data segments in the killer hill and steepest downhill segments are extracted. In the same relay, data from each segment is aggregated over all individuals to be used for group clustering and flow modeling. 

\section{Results}
\label{sec:Results}
%
%The performance of the proposed flow modeling and prediction algorithms are validated and compared in this section. 
%

%\subsection{Performance Evaluation}
%\label{subsec:PerfEval}
% 
%The flow modeling and prediction for single skier is first given in this section, followed by the modeling for multiple individuals in a cluster. 
%
%\begin{figure}
%	\centering
%	\includegraphics[width=8.5cm]{MSEvsS.eps}
%	\caption{MSE for SGP and on-line GP with various $s$.}
%	\label{fig:figure7}
%\end{figure}
%

\subsection{Force Analysis}
\label{subsec:forceAnalysisResults}
In this section, we investigate the relationship between performance and behavior of individual skiers. It is based on the force analysis introduced in Section~\ref{sec:ForceAnalysis}. The force analysis is done for both the killer hill and the steepest downhill segments of two tracks. 

\begin{figure}[tb]
	\centering
	\includegraphics[width=8.5cm,height=6.5cm]{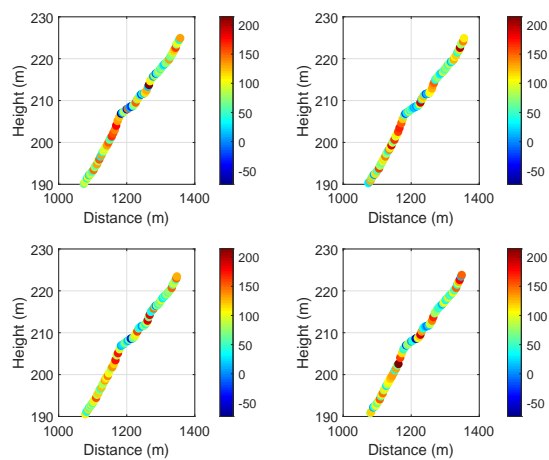}
	\caption{Resultant forces: Individual A, killer hill, track 1}
	\label{fig:fig3}
\end{figure}

\begin{figure}[tb]
	\centering
	\includegraphics[width=8.5cm,height=6.5cm]{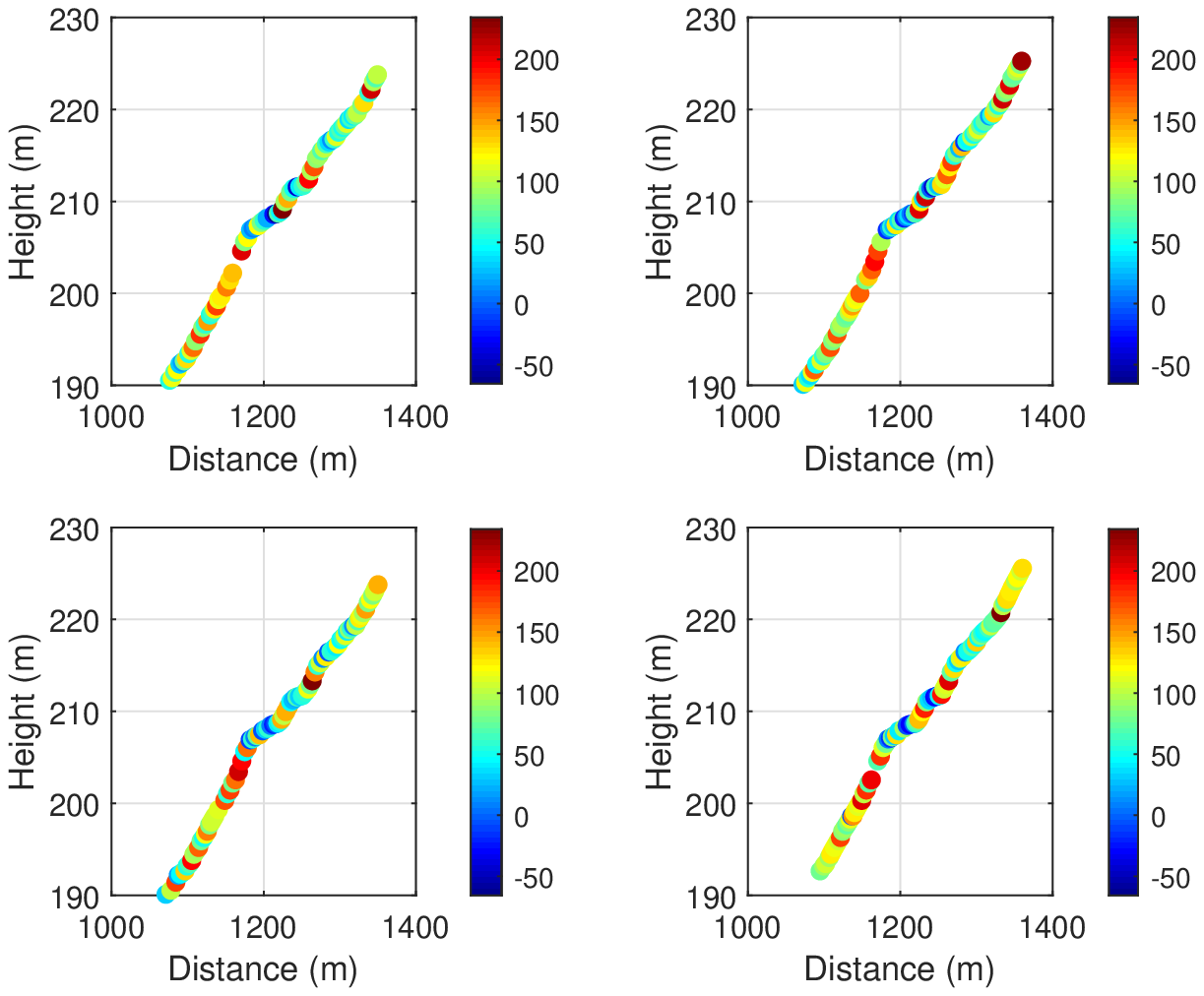}
	\caption{Resultant forces: Individual B, killer hill, track 1}
	\label{fig:fig4}
\end{figure}

\begin{figure}[tb]
	\centering
	\includegraphics[width=8.4cm,height=6.5cm]{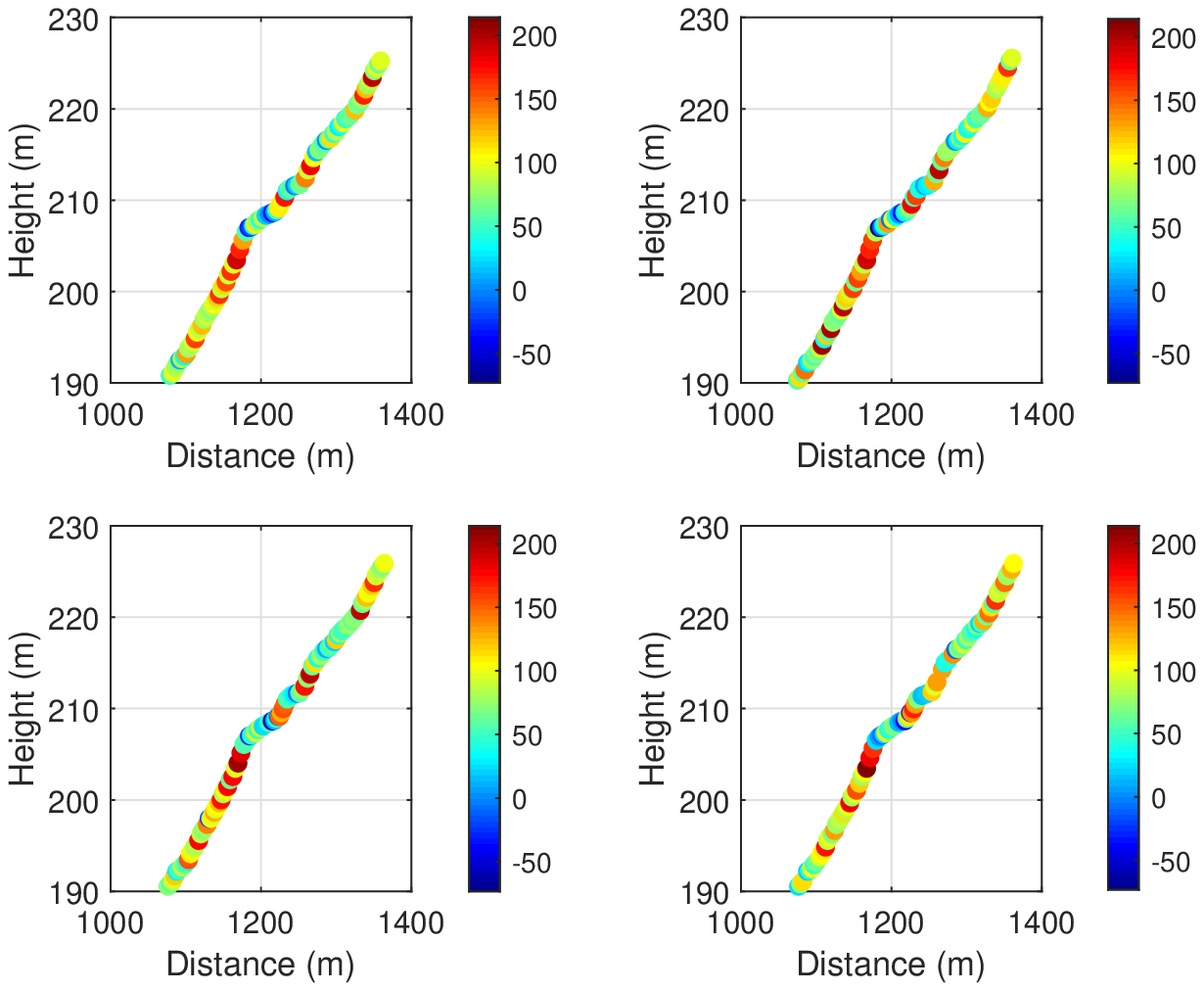}
	\caption{Resultant forces: Individual C, killer hill, track 1}
	\label{fig:fig5}
\end{figure}

\begin{figure}[tb]
	\centering
	\includegraphics[width=8.5cm,height=6.5cm]{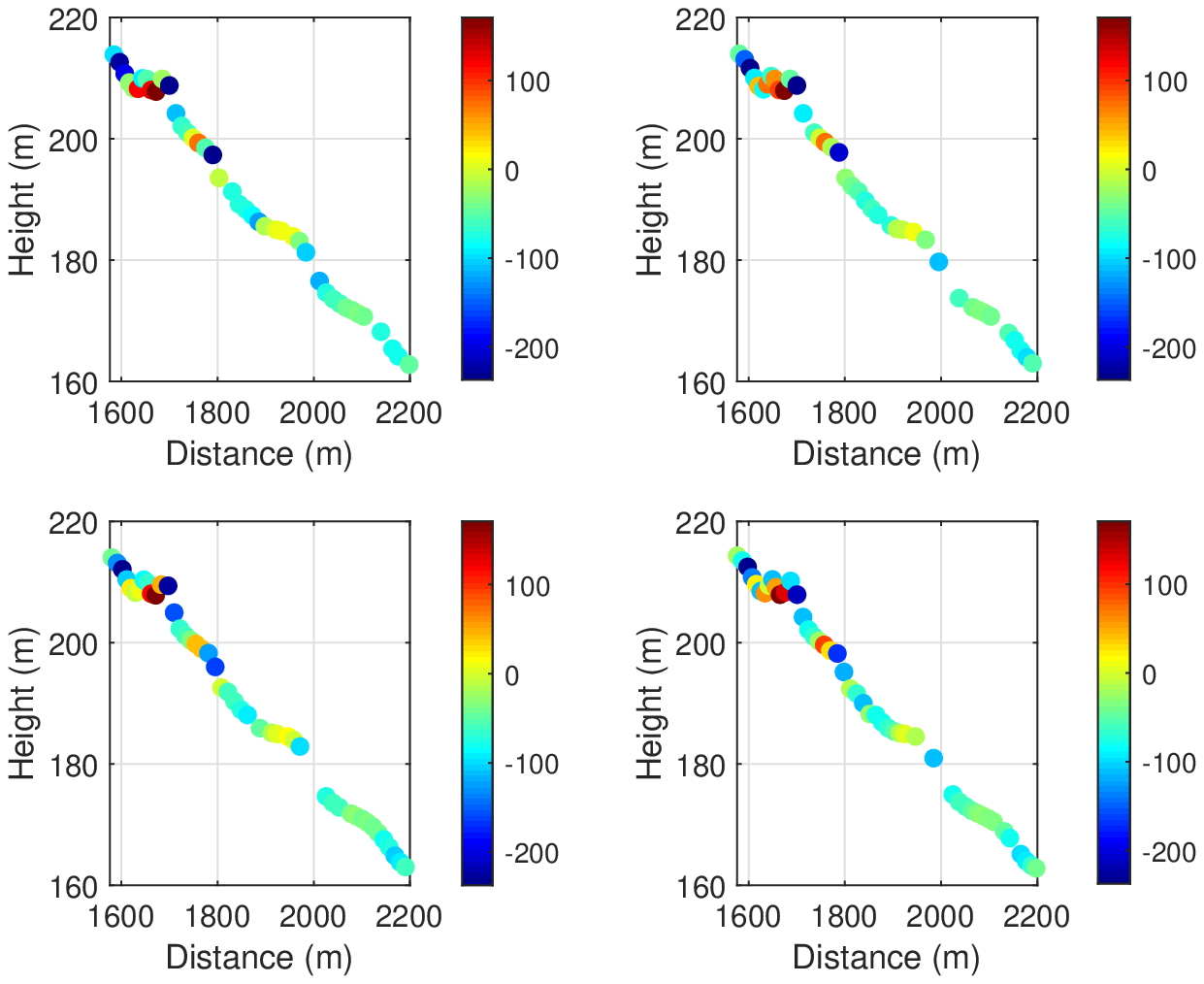}
	\caption{Resultant forces: Individual A, steepest downhill, track 1}
	\label{fig:fig6}
\end{figure}

\begin{figure}[tb]
	\centering
	\includegraphics[width=8.5cm,height=6.5cm]{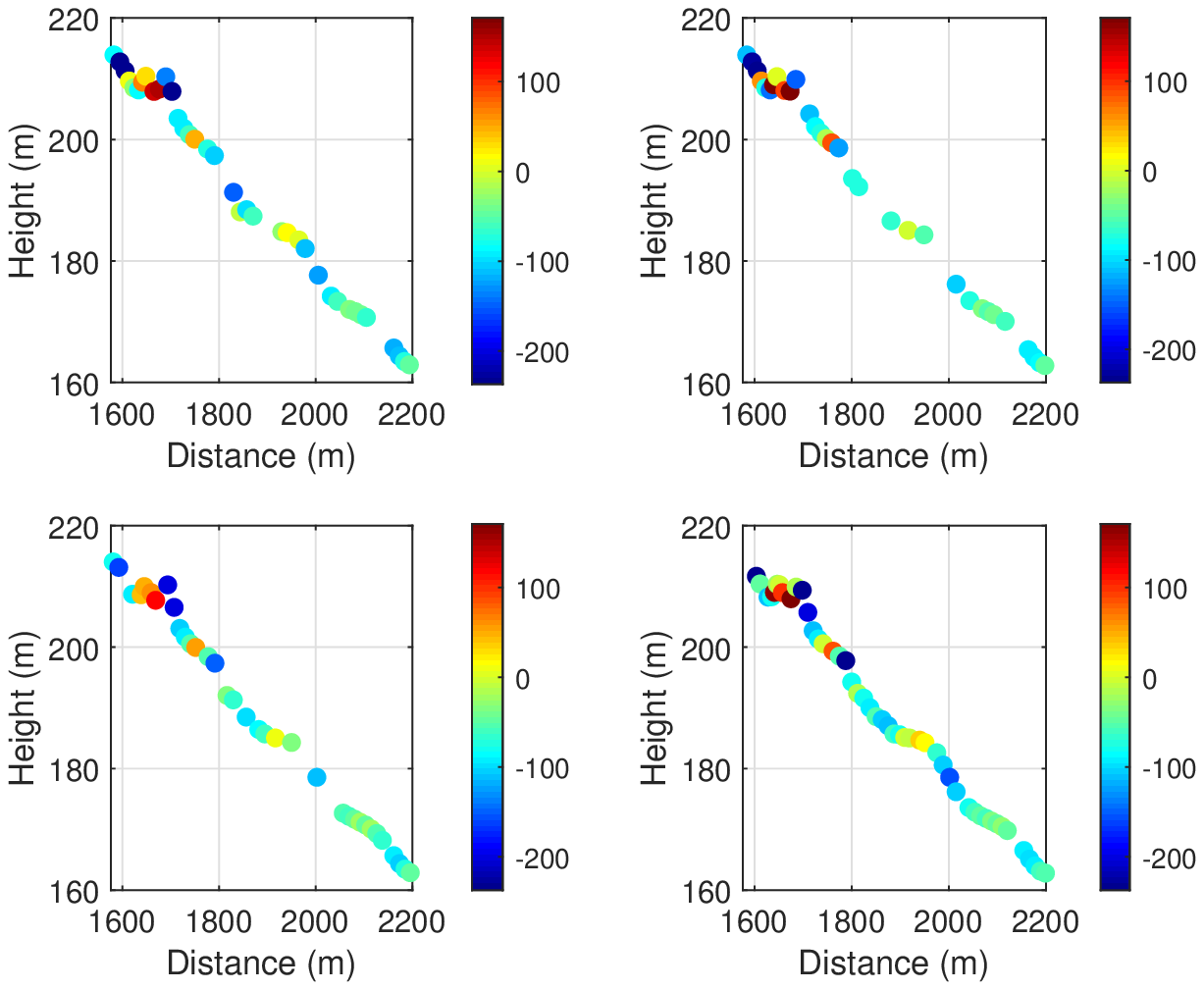}
	\caption{Resultant forces: Individual B, steepest downhill, track 1}
	\label{fig:fig7}
\end{figure}

\begin{figure}[tb]
	\centering
	\includegraphics[width=8.4cm,height=6.5cm]{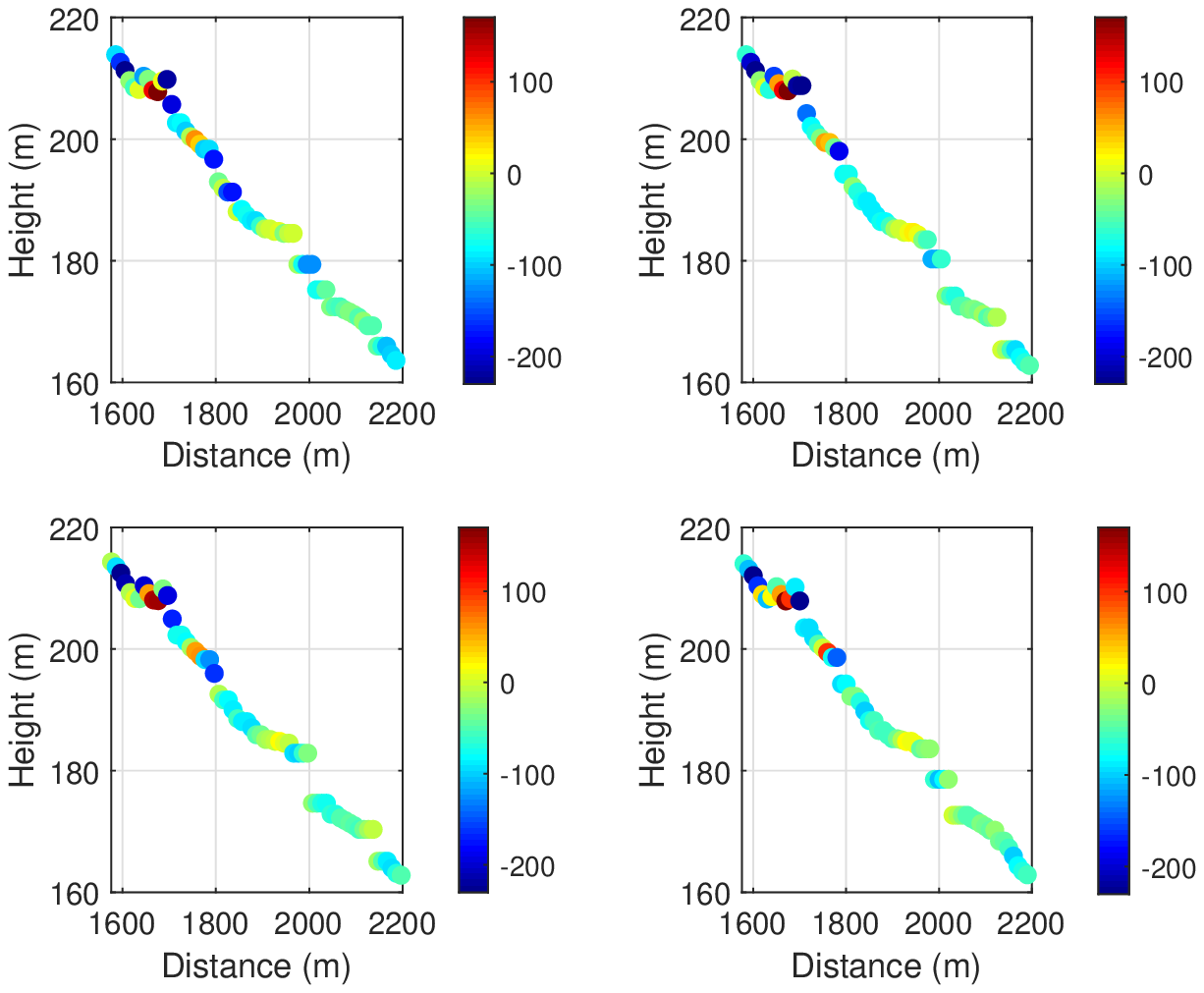}
	\caption{Resultant forces: Individual C, steepest downhill, track 1}
	\label{fig:fig8}
\end{figure}

First of all, the estimated forces on track $1$, the killer hill segment, are plotted in Fig.~\ref{fig:fig3} to \ref{fig:fig5}, with each figure representing i.e., one skier. Fig.~\ref{fig:fig6} to \ref{fig:fig8} depict the estimated forces at the steepest downhill segment. It is noted that for the steepest downhill segment, most parts are declining areas, while there are some small areas that are either inclining or rather flat. In addition, positive forces in the plots indicate resultant forces are acting in alignment with the moving direction, while the negative ones indicate the resultant forces are acting opposite to the moving direction. In total, we compare the forces on both the killer hill and steepest downhill for three different skiers, namely, individual A (best performance), B (competing with individual A) and C (fell behind).   

\begin{figure}[tb]
	\centering
	\includegraphics[width=8.4cm]{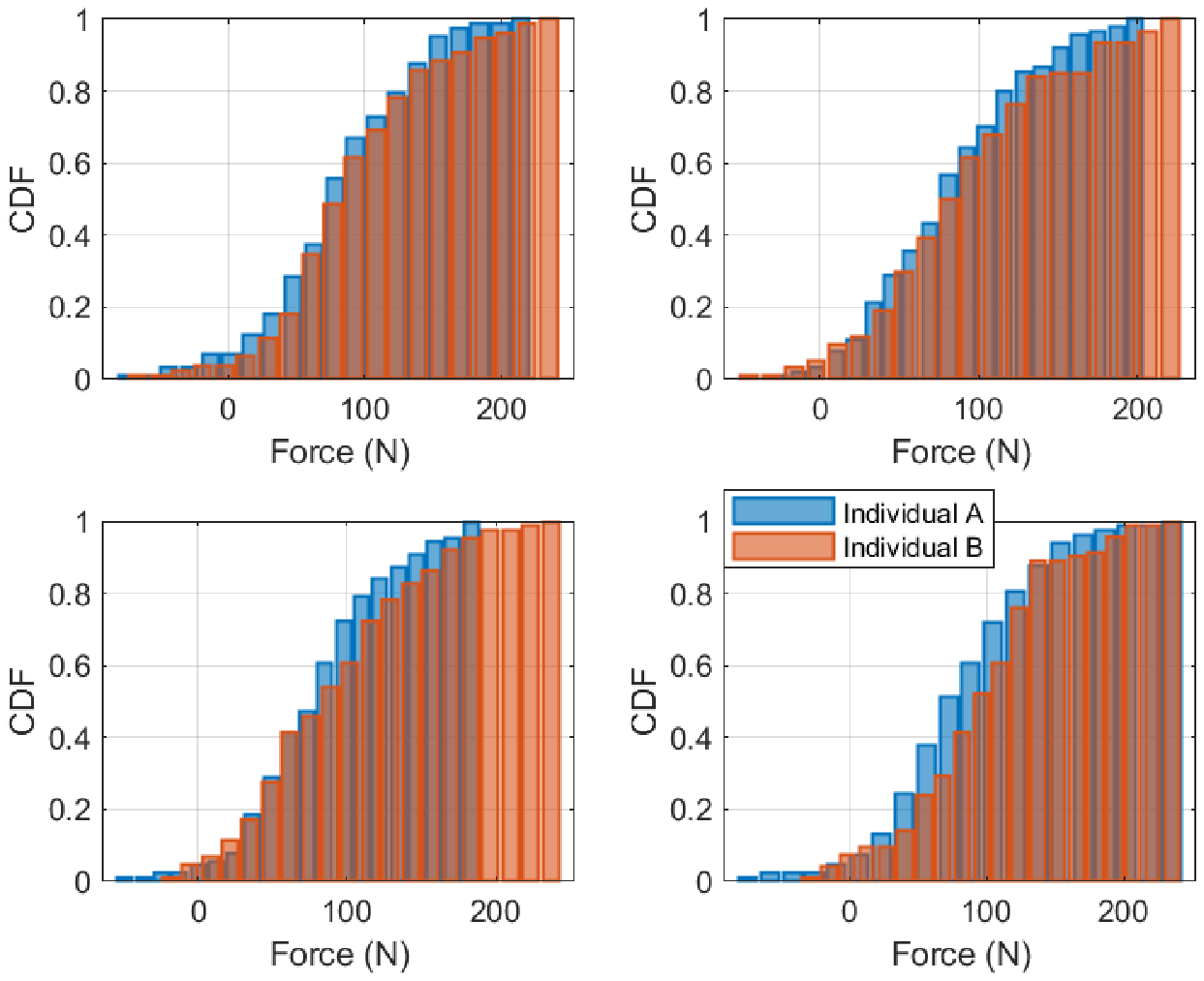}
	\caption{CDF of resultant forces: killer hill, track 1}
	\label{fig:fig15}
\end{figure}

\begin{figure}[tb]
	\centering
	\includegraphics[width=8.4cm]{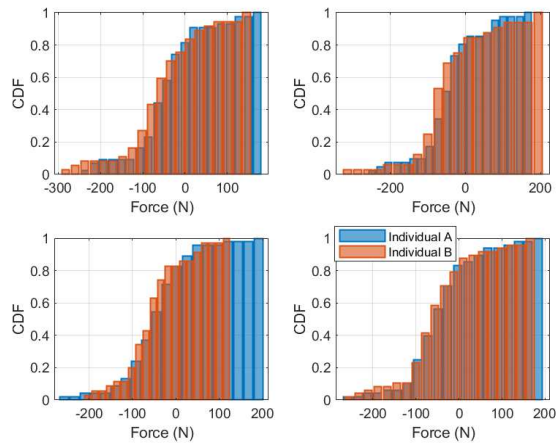}
	\caption{CDF of resultant forces: steepest downhill, track 1}
	\label{fig:fig16}
\end{figure}

From all the plots for track $1$, where classic style is applied, we have the following observations: (I) Skier A and B have stronger forces on average at the killer hill segment and hence outperform skier C;(II) Larger forces/frictions are estimated at sharper slopes, for instance, between $1100$ to $1200$ meters at killer hill and around $1600$ and $1700$ meters at steepest downhill;(III) Different strategies were applied, for instance, skier A has larger forces in lap 1 and 4, where skier B has larger forces in lap 2 and 3, and skier C has almost evenly distributed forces for all 4 laps. To further verify the observations, histograms of the estimated resultant forces for two segments are shown in Fig.~\ref{fig:fig15} and \ref{fig:fig16}. Due to space limitation, we only show the comparison between individual A and B, who were competing with each other. It can be observed that for the killer hill, skier A and B distribute forces differently over 4 laps. At the steepest down hill, individual B experience larger negative forces, especially for lap 1, 2 and 4. It should be noted that individual B has much larger weight than individual A, which may result in higher friction when declining.  

\begin{figure}[tb]
	\centering
	\includegraphics[width=8.5cm,height=6.5cm]{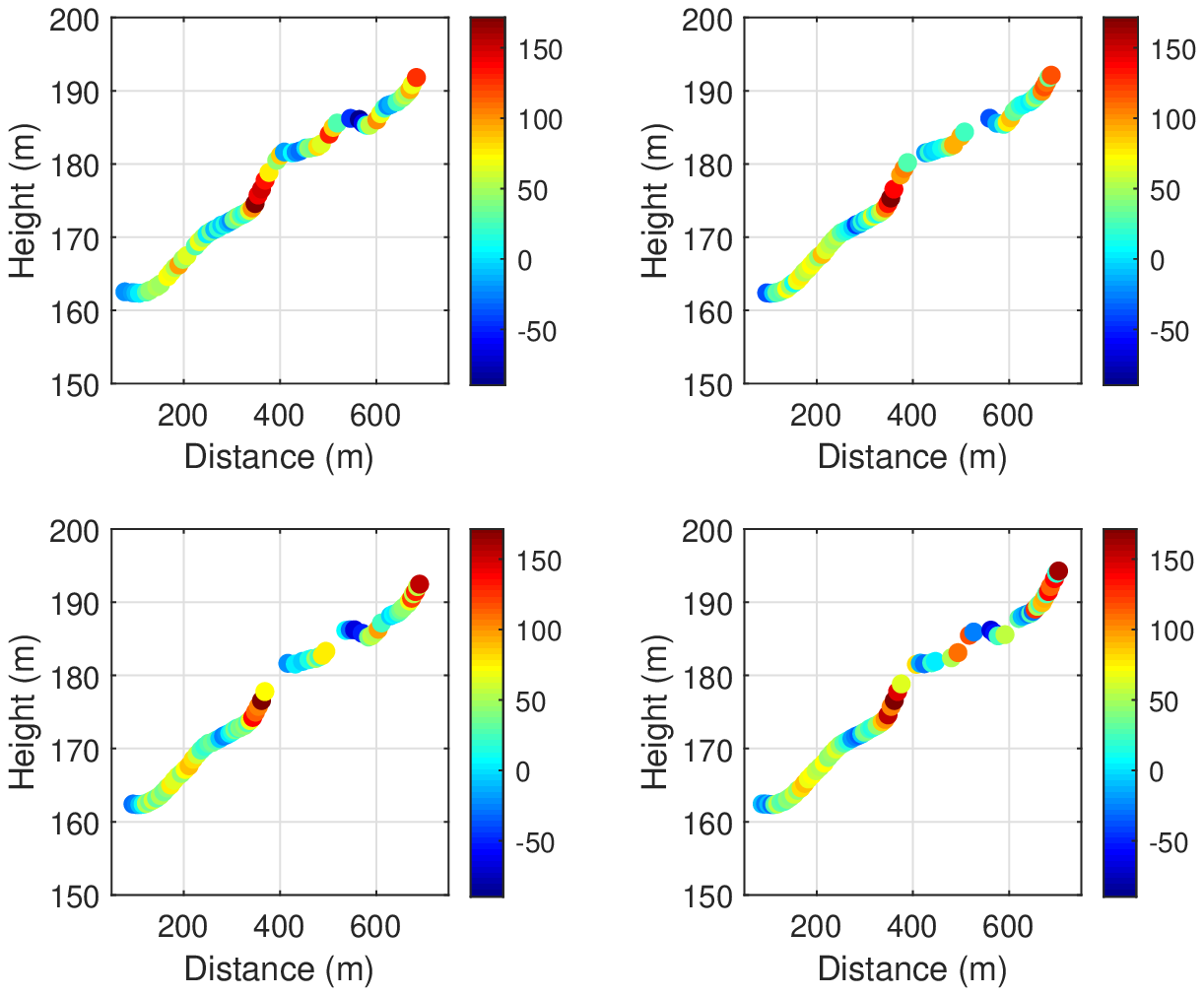}
	\caption{Resultant forces: Individual D, killer hill, track 2}
	\label{fig:fig9}
\end{figure}

\begin{figure}[tb]
	\centering
	\includegraphics[width=8.5cm,height=6.5cm]{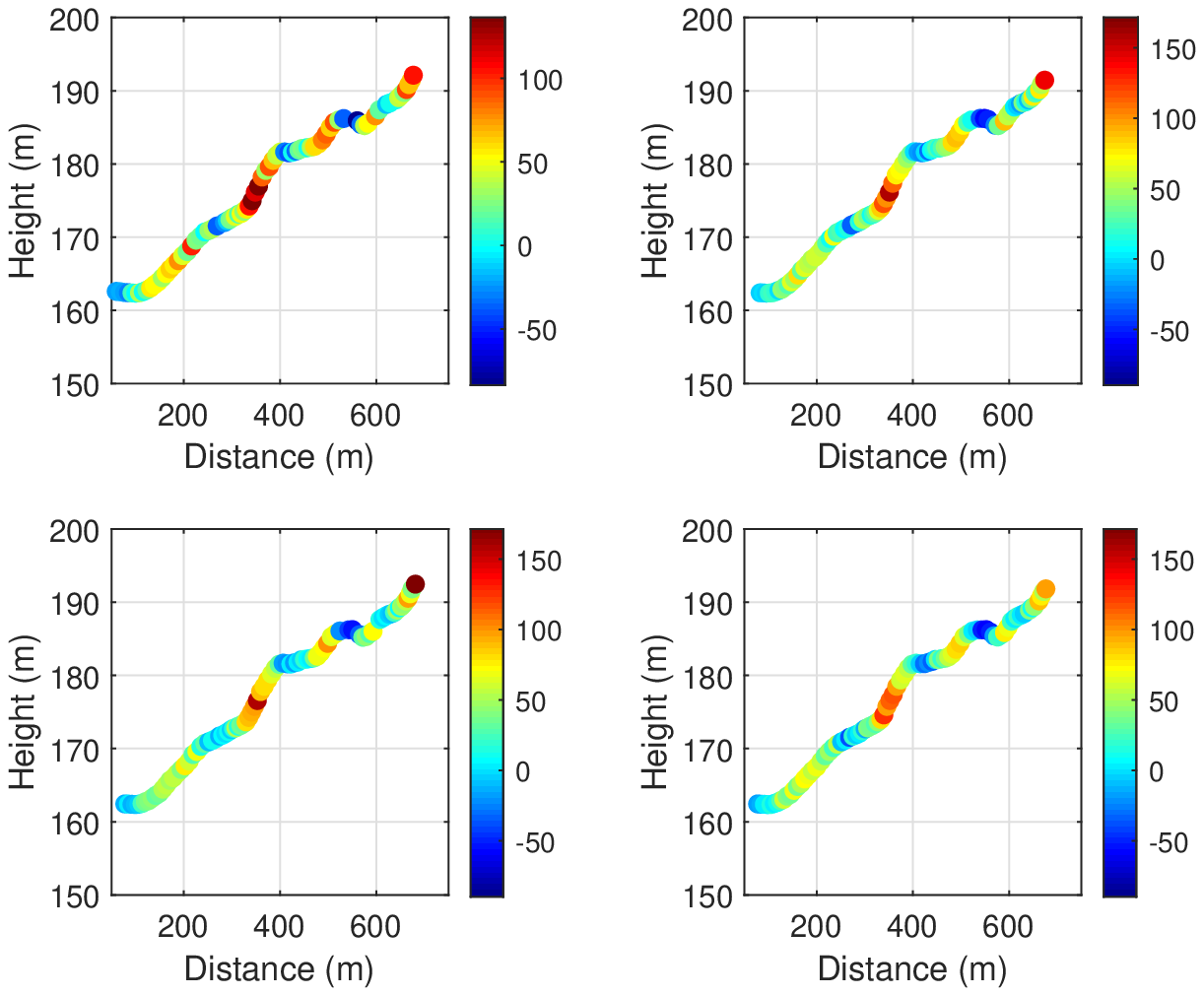}
	\caption{Resultant forces: Individual E, killer hill, track 2}
	\label{fig:fig10}
\end{figure}

\begin{figure}[tb]
	\centering
	\includegraphics[width=8.4cm,height=6.5cm]{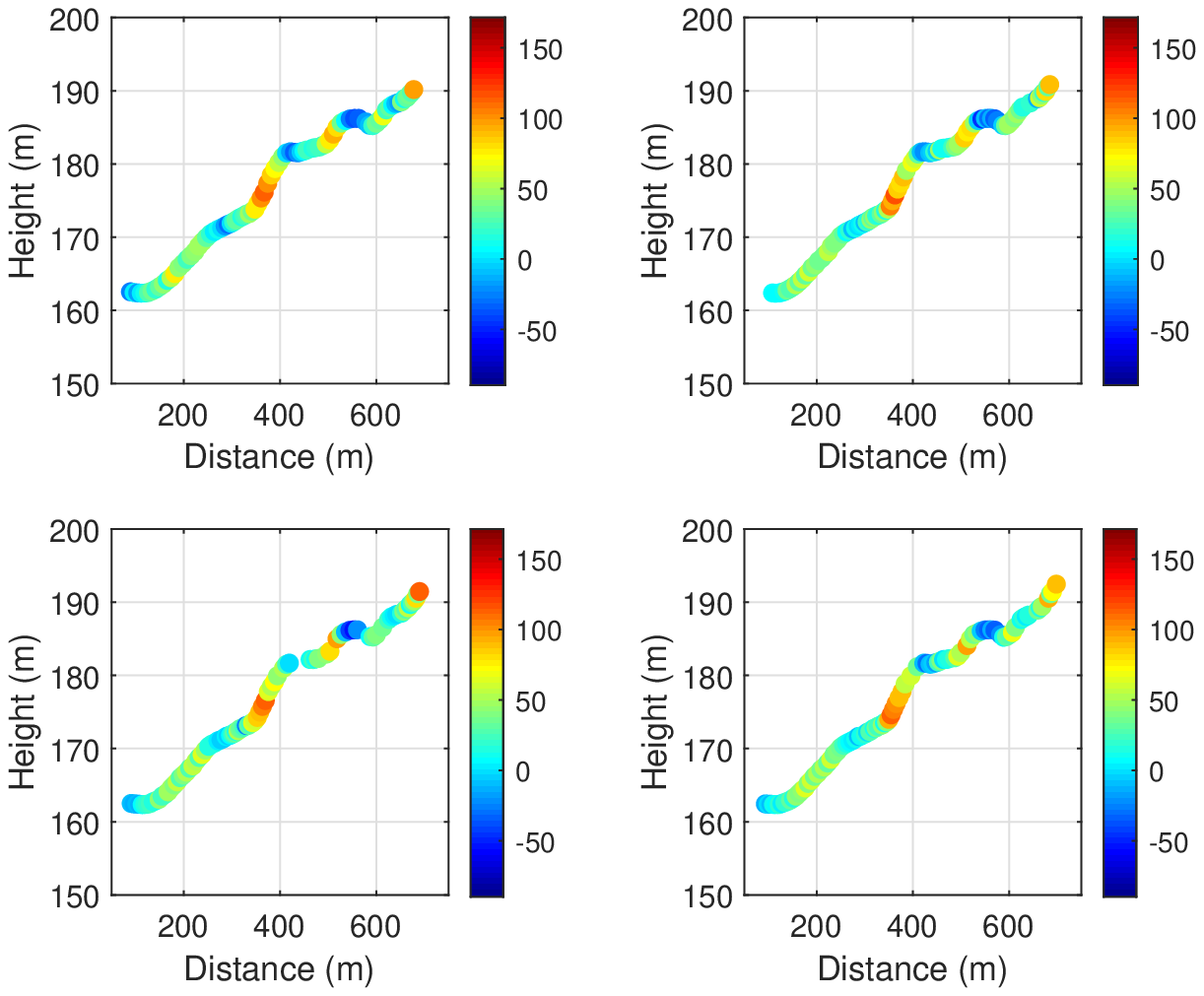}
	\caption{Resultant forces: Individual F, killer hill, track 2}
	\label{fig:fig11}
\end{figure}

\begin{figure}[tb]
	\centering
	\includegraphics[width=8.5cm,height=6.5cm]{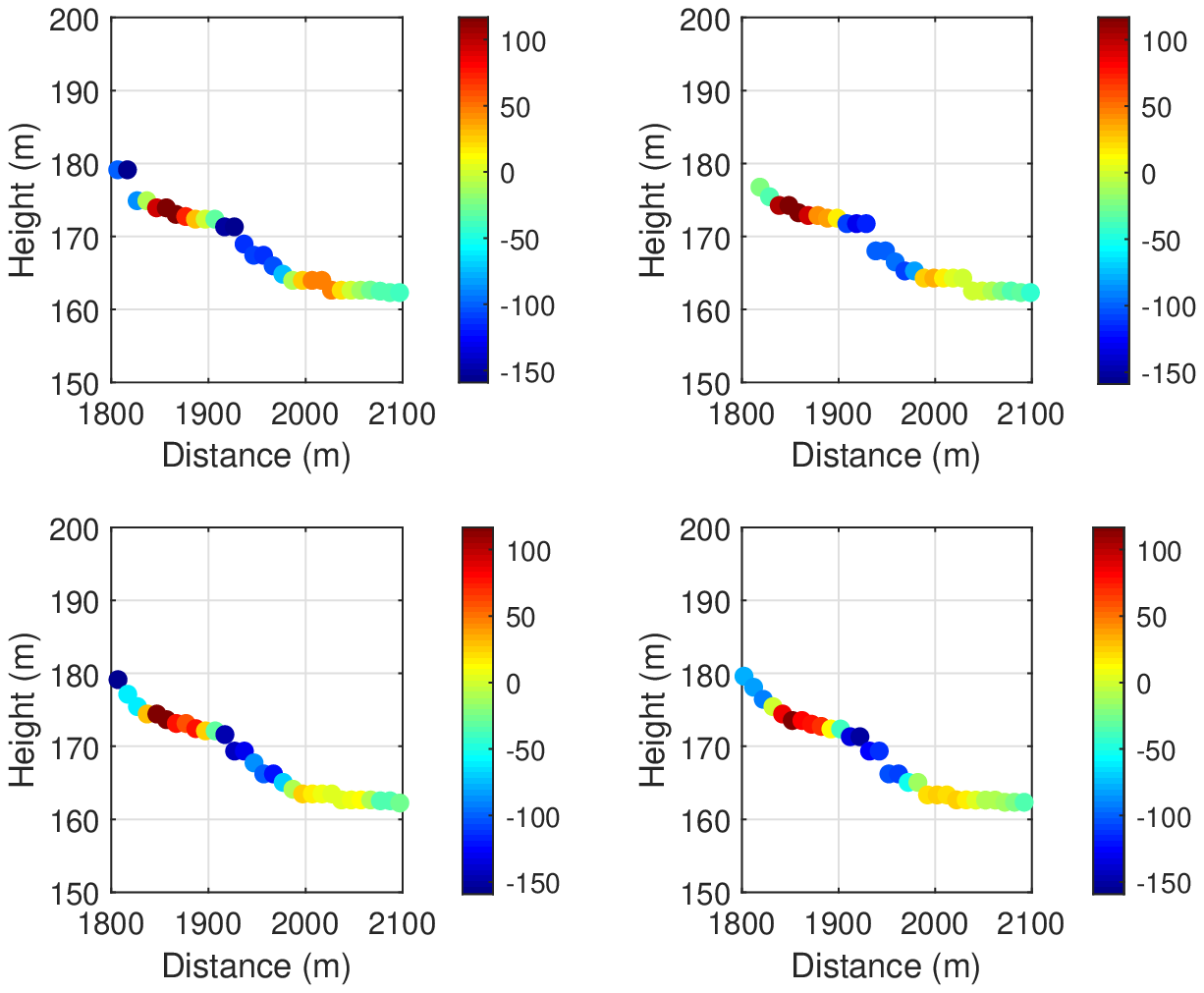}
	\caption{Resultant forces: Individual D, steepest downhill, track 2}
	\label{fig:fig12}
\end{figure}

\begin{figure}[tb]
	\centering
	\includegraphics[width=8.5cm,height=6.5cm]{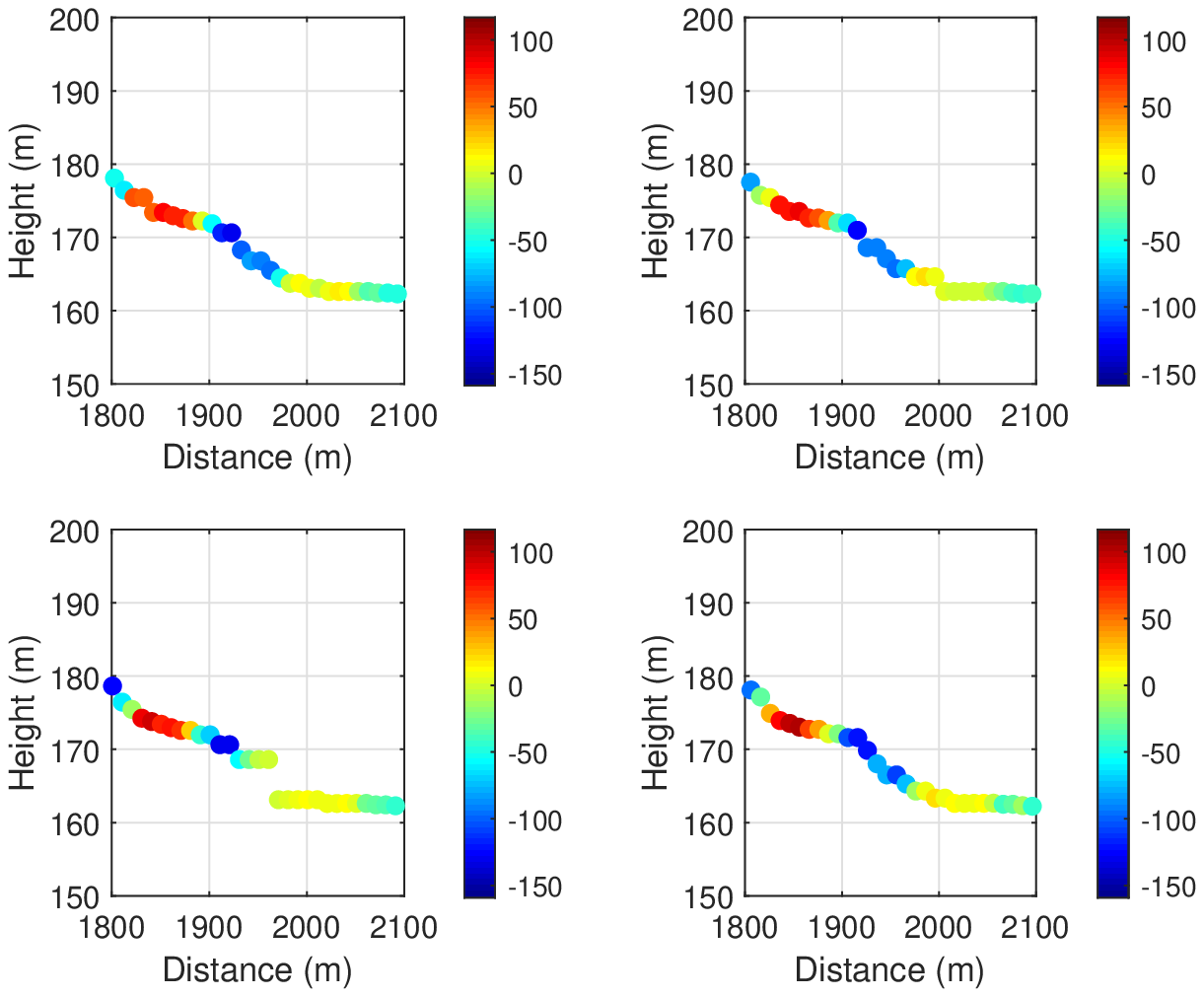}
	\caption{Resultant forces: Individual E, steepest downhill, track 2}
	\label{fig:fig13}
\end{figure}

\begin{figure}[tb]
	\centering
	\includegraphics[width=8.4cm,height=6.5cm]{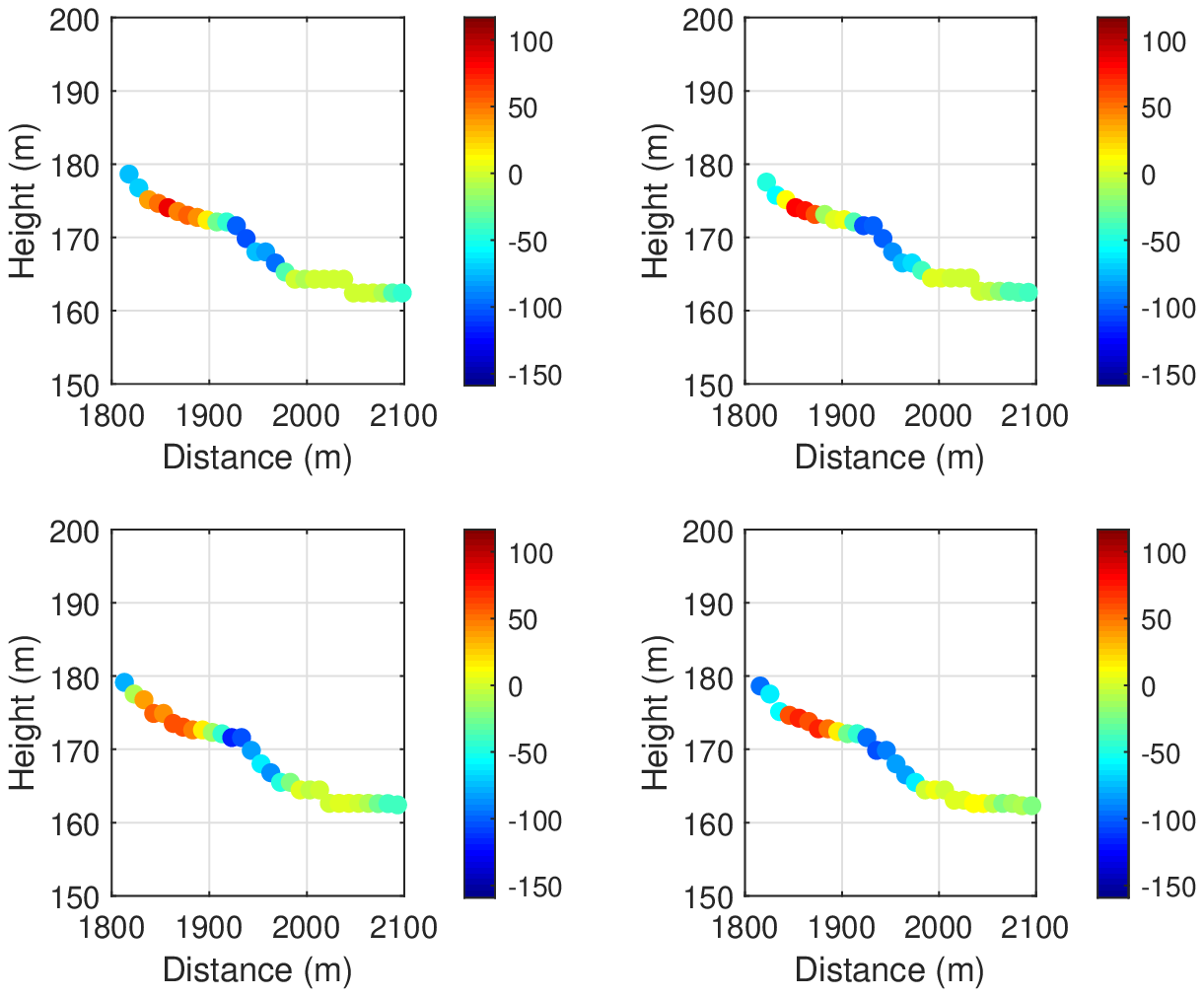}
	\caption{Resultant forces: Individual F, steepest downhill, track 2}
	\label{fig:fig14}
\end{figure}

\begin{figure}[tb]
	\centering
	\includegraphics[width=8.4cm]{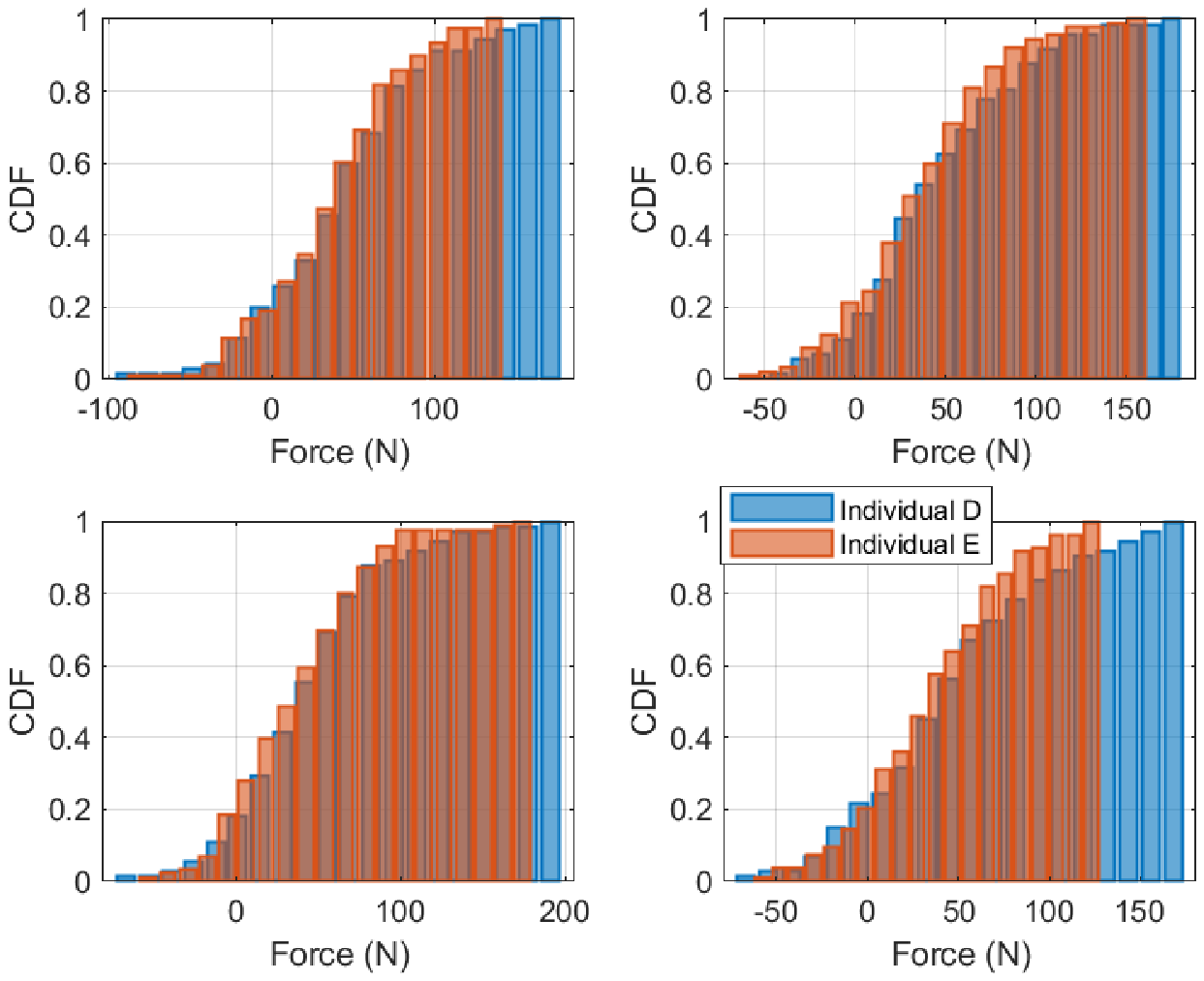}
	\caption{CDF of resultant forces: killer hill, track 2}
	\label{fig:fig17}
\end{figure}

\begin{figure}[tb]
	\centering
	\includegraphics[width=8.4cm]{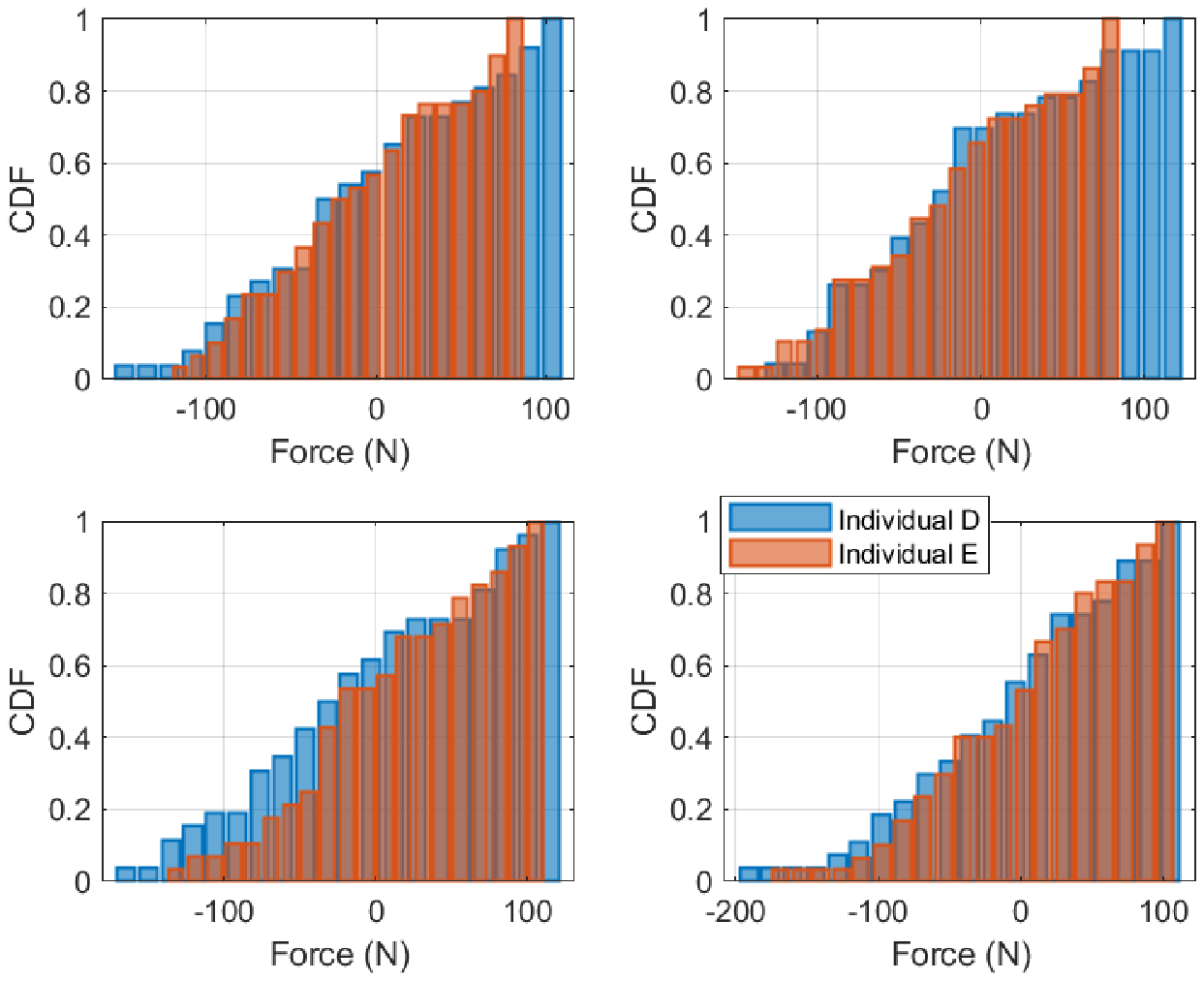}
	\caption{CDF of resultant forces: steepest downhill, track 2}
	\label{fig:fig18}
\end{figure}

Similarly, the estimated resultant forces are evaluated on track $2$, which are illustrated in Fig.~\ref{fig:fig9} to \ref{fig:fig14} for both a killer hill and a steepest downhill segment, for individual D (best performance), E (competing with individual D), F (fell behind), respectively. The histogram of individual D and E are also compared for both segments in Fig.~\ref{fig:fig17} and \ref{fig:fig18}. It can be observed that for the killer hill, individual D has larger positive forces than E (especially in lap 4, where individual D performed an overtaking of individual E). It is also noted that the higher the weight is (e.g., individual D is heavier than E, and individual F is the lightest), the larger friction the individual will experience when declining (e.g, see Fig.~\ref{fig:fig12} to \ref{fig:fig14} between $1900$ and $2000$ meters). Compared with forces on track $1$, where the classic style is applied, the forces are much smaller on track $2$, this is probably due to the fact that different skiing technique (i.e., skating style) is used. It is also worth mention that for the steepest downhill segment, the forces almost have a uniform distribution between $-100$ and $100$ Newton on track $2$.   

\subsection{Individual Flow Model and Prediction}
\label{sub:IndResults}
\begin{figure}[tb]
	\centering
	\includegraphics[width=8.5cm]{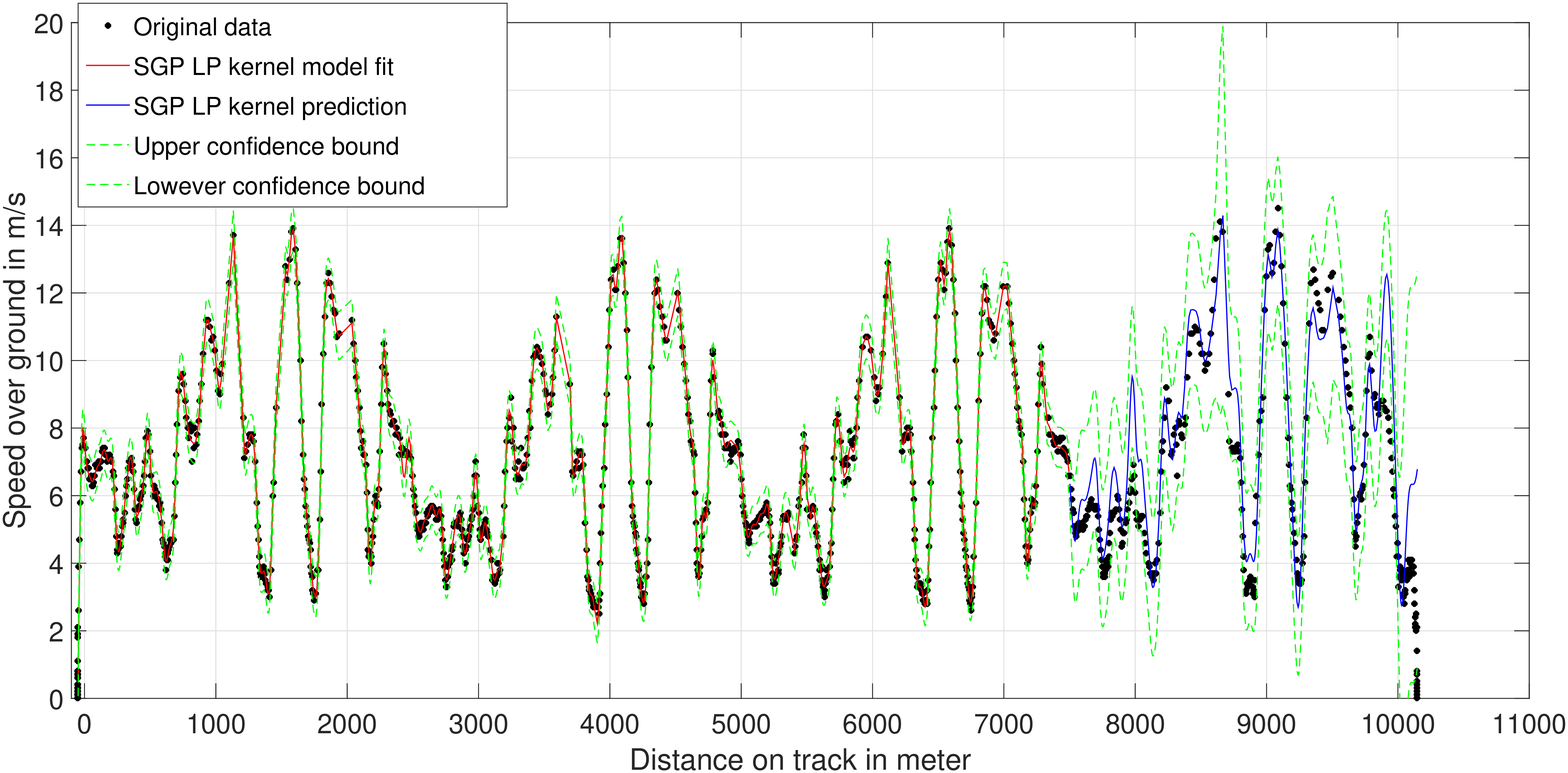}
	\caption{Flow model (first $3$ laps) and prediction (lap $4$) for SGP: LP kernel.}
	\label{fig:figure4}
\end{figure}

\begin{figure}[tb]
	\centering
	\includegraphics[width=8.5cm]{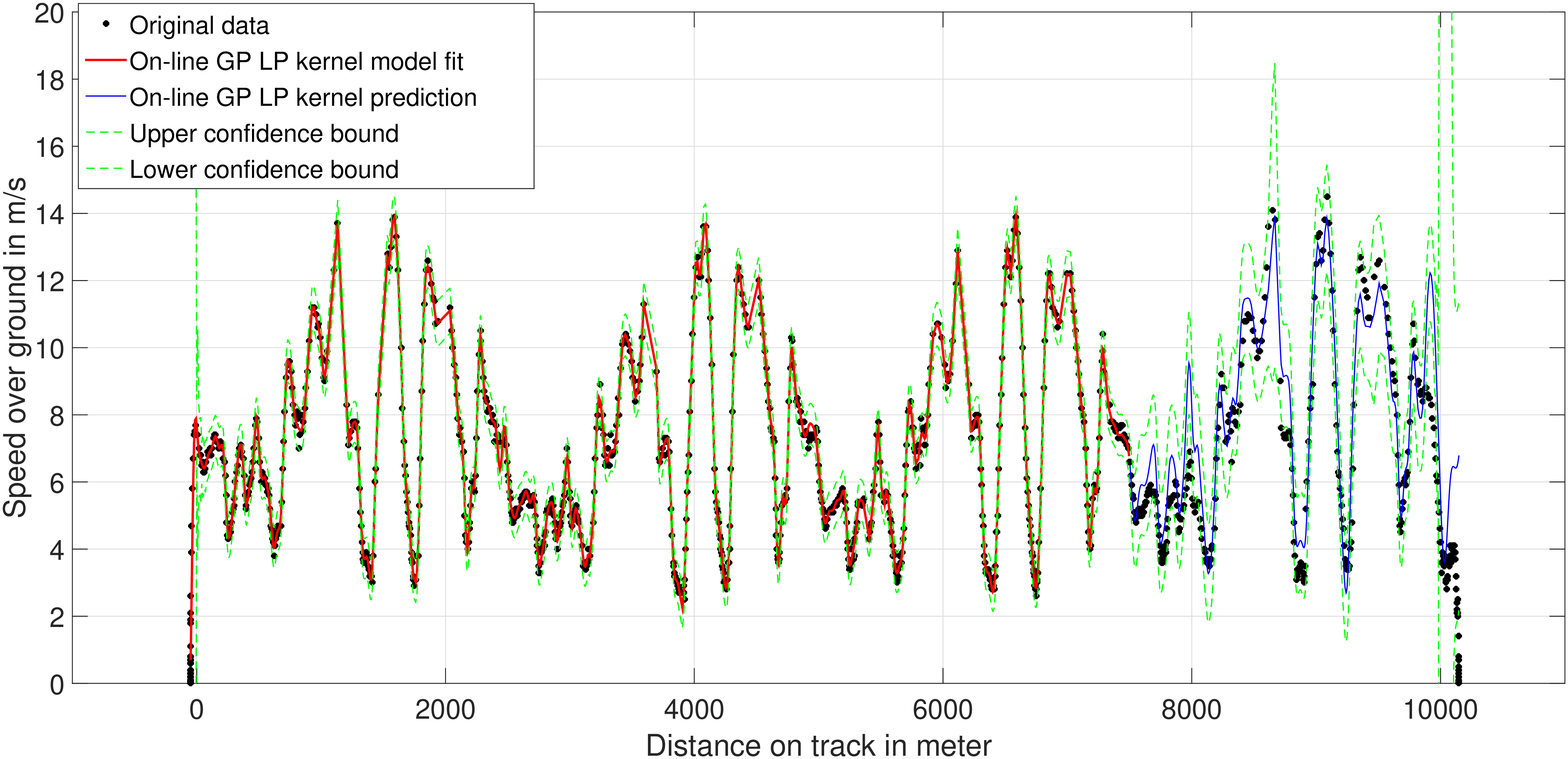}
	\caption{Flow model (first $3$ laps) and prediction (lap $4$) for OGP: LP kernel.}
	\label{fig:figure6}
\end{figure}

 The ground speed versus distance on track for one specific individual is depicted in Fig. \ref{fig:figure4}. The speed in four laps shows a periodic pattern, while difference can also be observed between different laps. For instance, better performance in lap 4 has been observed, when the individual sprint for the last lap. In order to evaluate the goodness of fit and compare the predictions made by different models, the data from the first three laps are used for training and the data from the last lap is used for validation. The results for the SGP and the OGP with different kernels have been shown in Fig. \ref{fig:figure4} and \ref{fig:figure6}, respectively. For the OGP in Fig. \ref{fig:figure6}, $s = 500$ grid points are uniformly selected within the race distance, i.e., 10 kilometers.  %For SGP, two kernels, namely the SE kernel and the LP kernel, are evaluated and compared. Note that in Fig. \ref{fig:figure5}, only the upper $95\%$ confidence bound is plotted for illustration, the lower bound is symmetric w.r.t the posterior mean (mostly equal to zero).For each individual, we model the ground speed given a distance on track (i.e., uniquely corresponds to a 2-D geographical position on the track) using both the standard and the grid based OGP.
%\begin{figure}[tb]
%	\centering
%	\includegraphics[width=8.5cm]{Figure22.eps}
%	\caption{Flow model (first $3$ laps) and prediction (lap $4$) for SGP: SE kernel.}
%	\label{fig:figure5}
%\end{figure}
%\begin{figure}[tb]
%	\centering
%	\includegraphics[width=8.5cm]{Figure24.eps}
%	\caption{MSE and processing time in MATLAB for SGP and OGP with various $s$.}
%	\label{fig:figure8}
%\end{figure}

 From Fig. \ref{fig:figure4} and \ref{fig:figure6}, we can see that both GPs provide good fit for the training data. Both SGP and OGP give good performance in prediction using the LP kernel. 
%, while bad performance using the SE kernel. The prediction performance greatly degrades as the training data runs out since the posterior mean converges rapidly to the prior mean without new data coming in
We can also see that the OGP with $s=500$ shows similar performance as the SGP with training data of size  $M=1219$. For further comparisons between the two GPs, see for instance \cite{YFFFJ16}. The mean predictive standard deviation for the speed difference between two consecutive time instants (i.e., $\Delta v=v_t-v_{t-1}$) is compared in Table~\ref{tab:1}, where the comparison is for skier E, in the killer hill segments for all 4 laps. For the black-box approach for individual skier, we compute the predictive variance for $v_t$ and $v_{t-1}$, namely, $\hat{\sigma}^2(d_t)$, and $\hat{\sigma}^2(d_{t-1})$ according to \eqref{eq:predictedCov}. The predictive variance for the speed difference is computed as $\sigma^2_{\Delta v}=\hat{\sigma}^2(d_t)+\hat{\sigma}^2(d_{t-1})$. The average standard derivations for the killer hill segment is then computed by taking the mean value of $\sigma_{\Delta v}$ during the killer hill time interval. For the grey-box approach, the predictive variance for can be computed from $\hat{\sigma}_r^2(d_t)$ and the force model given in \eqref{eq:kineticGP}, yielding
\begin{equation}
\sigma^2_{\Delta v}=\frac{\hat{\sigma}_r^2(d_t)}{m^2}\Delta t^2.
\end{equation}
Similarly, the mean of the standard deviation $\sigma_{\Delta v}$ during the killer hill segment is registered in Table~\ref{tab:1}. From the comparison, we have observed larger predictive standard deviation for the  black-box approach. This may due to the fact that in black-box approach, the model is completely unknown, which leads to larger ambiguity in the prediction. %the mean-squared-error (MSE) and the processing time for the two GPs are illustrated in Fig. \ref{fig:figure8}. The MSE is computed from a $4$-fold cross-validation \citep{Bishop06}, and each fold is obtained by taking the data from a corresponding lap. In each iteration, one fold is left out for validation and the rest are used for training. 
%\begin{table}[t]
%	\scriptsize
%	\centering
%	\caption{Comparison between black-box and grey-box approach}
%	\setlength\tabcolsep{3pt} 
%	\begin{tabular}[htdp]{llll}
%		\hline
%		\textbf{Lap number}                   & \textbf{Black-box} &\textbf{Grey-box}\\
%		\hline
%		$1$                 & & \\
%		%		\hline
%		$2$         & & \\
%		%		\hline  
%		$3$      & &  \\
%		$4$ & & \\
%		\hline
%	\end{tabular}
%	\label{tab:cases}
%\end{table}
\begin{table}
	\caption{Comparison between black-box and grey-box approach: the mean predictive standard derivation for the speed difference $\Delta v=v_t-v_{t-1}$}.
	\label{tab:1}       % Give a unique label
	\begin{tabular}{lll}
		\hline\noalign{\smallskip}
		Lap & \textbf{Black-box (m/s)} &\textbf{Grey-box (m/s)} \\
		\noalign{\smallskip}\hline\noalign{\smallskip}
%		1 & 0.1721 & 0.0913 \\
%		2 & 0.1694 & 0.0592 \\
%		3 & 0.1767 & 0.0940 \\
%		4 & 0.1656 & 0.0487 \\
1 & 0.4148 & 0.3022 \\
2 & 0.4116 & 0.2433 \\
3 & 0.4204 & 0.3066 \\
4 & 0.4069 & 0.2207 \\
		\noalign{\smallskip}\hline
	\end{tabular}
\end{table}
%Seen from Fig. \ref{fig:figure8}, the processing time (i.g., excluding the parameter training time) for OGP is greatly reduced. Compared with SGP, even lower MSE for prediction is achieved for OGP with $s \geq 100$. For OGP, the posterior mean and variance of the predefined grid points are updated as training data coming in sequentially. If the grid points are properly defined, when there comes a new distance value, the grid points close to it will provide more information. However, in the SGP, the prediction at a new distance will depend on the batch training data which may be far away from it. Hence with a proper selected $s$, OGP provides a better prediction accuracy than SGP. 
%
\subsection{Aggregated Flow Modeling (Multiple Individuals)}
\label{subsec: AggResults}
Clustering is performed first for individuals on the same track in the same relay. Here we focus on the killer hill and steepest downhill segments as shown in Figure \ref{fig:figure1} where the individuals may perform differently. 
%
%
%The highlighted part of Fig.~\ref{fig:figure1} is a steep uphill where the individuals may perform quite differently. 
Clustering of the ground speed curves of two individuals is similar to clustering of line-of-sight and non-line-of-sight signal waveforms described in \citep{WMG12}. The features considered in this work are: maximum, minimum, variance and mean value of speed, energy, skewness and kurtosis. With the features extracted from the corresponding data segments, clustering is performed as described in \cite[Section V.B]{YFFFJ16}. The number of clusters is $N_k = 2$ in this evaluation.

%repeatedly for each individual as follows:
%%
%\begin{enumerate}
%	\item Segment the complete dataset of each individual and extract the segments with interested locations. 
%	\item Extract the features for each individual $\mathbf{f}\! \triangleq \![\!v_{max},  v_{min}, \bar{v}, v_{\sigma}, v_{\varepsilon}, v_{sk}, v_{kr}\!]\!^T$ from the selected segment. 
%	\item Set the number of clusters to $N_k$. For all the feature vectors $\mathbf{f}_1, \mathbf{f}_2, \ldots, \mathbf{f}_{N_s}$, where $N_s$ is the number of individuals, normalization is performed. Then, the canonical K-means algorithm \citep{Bishop06} is applied for clustering.
%	\item Store the cluster indicator for each individual.
%\end{enumerate}
 %For illustration purpose, we only show the clustering results for relay $1$, lap $4$ on track $1$ in Fig. \ref{fig:figure9} and \ref{fig:figure10}. 
%
%
%
%\begin{figure}[tb]
%	\centering
%	\includegraphics[width=8.5cm]{Figure25.eps}
%	\caption{Two clusters of individuals in lap $4$ track $1$: killer hill segment. Top: cluster $1$; bottom: cluster $2$}
%	\label{fig:figure9}
%\end{figure}
%%
%\begin{figure}[tb]
%	\centering
%	\includegraphics[width=8.5cm]{Figure26.eps}
%	\caption{Two clusters of individuals in lap $4$ track $1$: steepest downhill segment. Top: cluster $1$; bottom: cluster $2$}
%	\label{fig:figure10}
%\end{figure} 
\begin{figure}[tb]
	\centering
	\includegraphics[width=8.4cm]{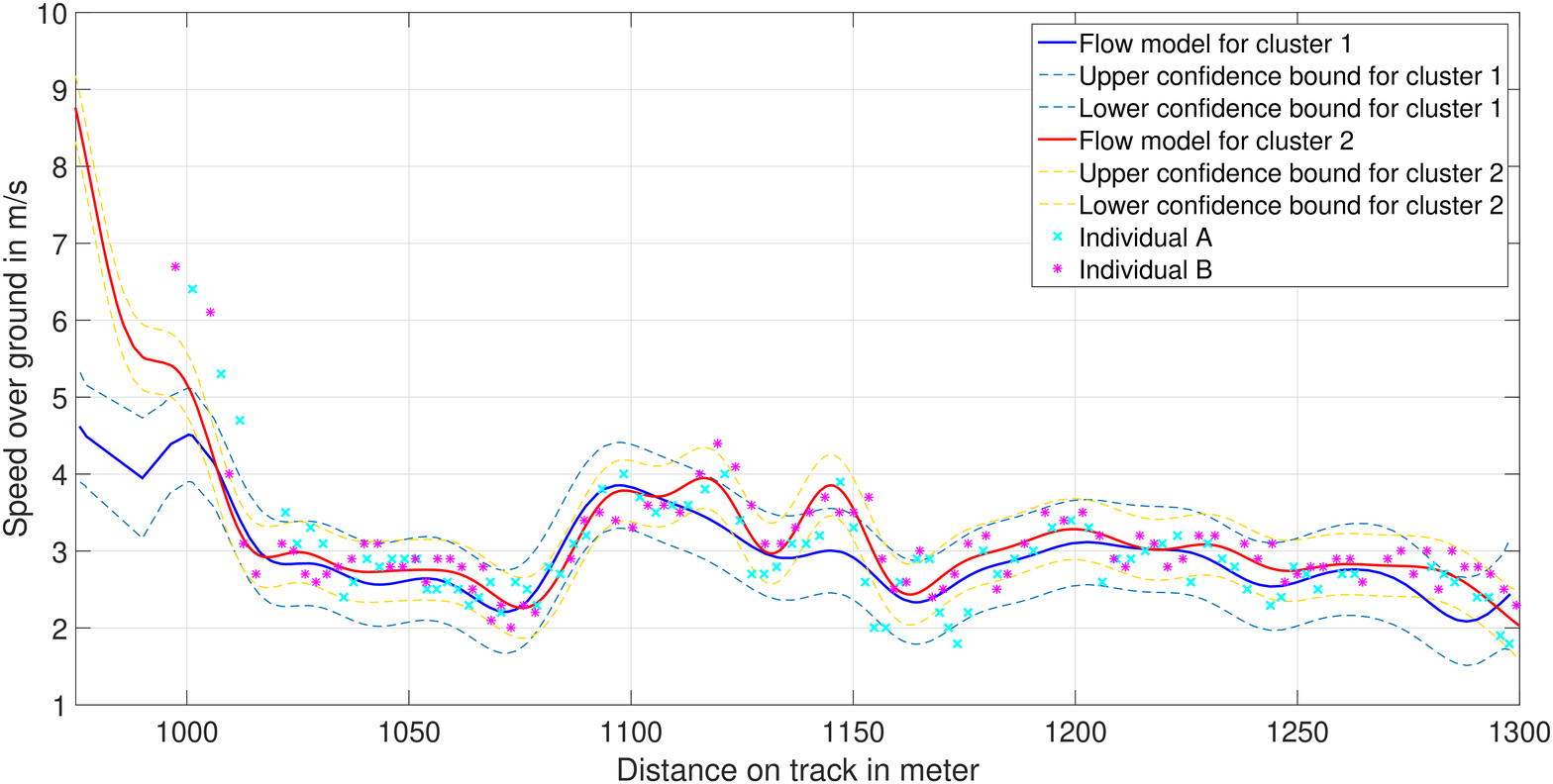}
	\caption{Flow models of individuals in lap $1$ track $1$: killer hill.}
	\label{fig:figure11}
\end{figure}
\begin{figure}[tb]
	\centering
	\includegraphics[width=8.5cm]{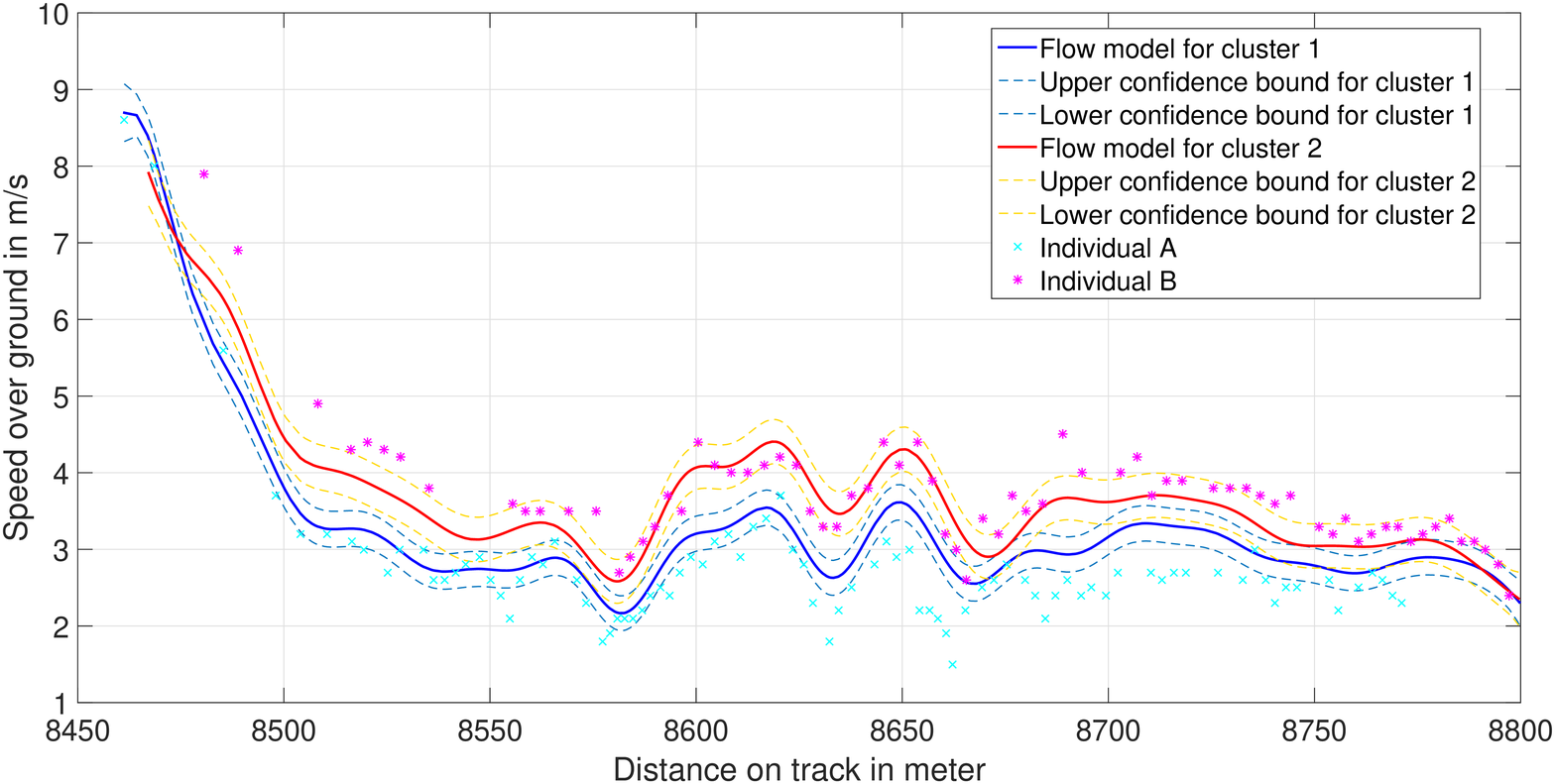}
	\caption{Flow models of individuals in lap $4$ track $1$: killer hill.}
	\label{fig:figure14}
\end{figure}
\begin{figure}[tb]
	\centering
	\includegraphics[width=8.7cm]{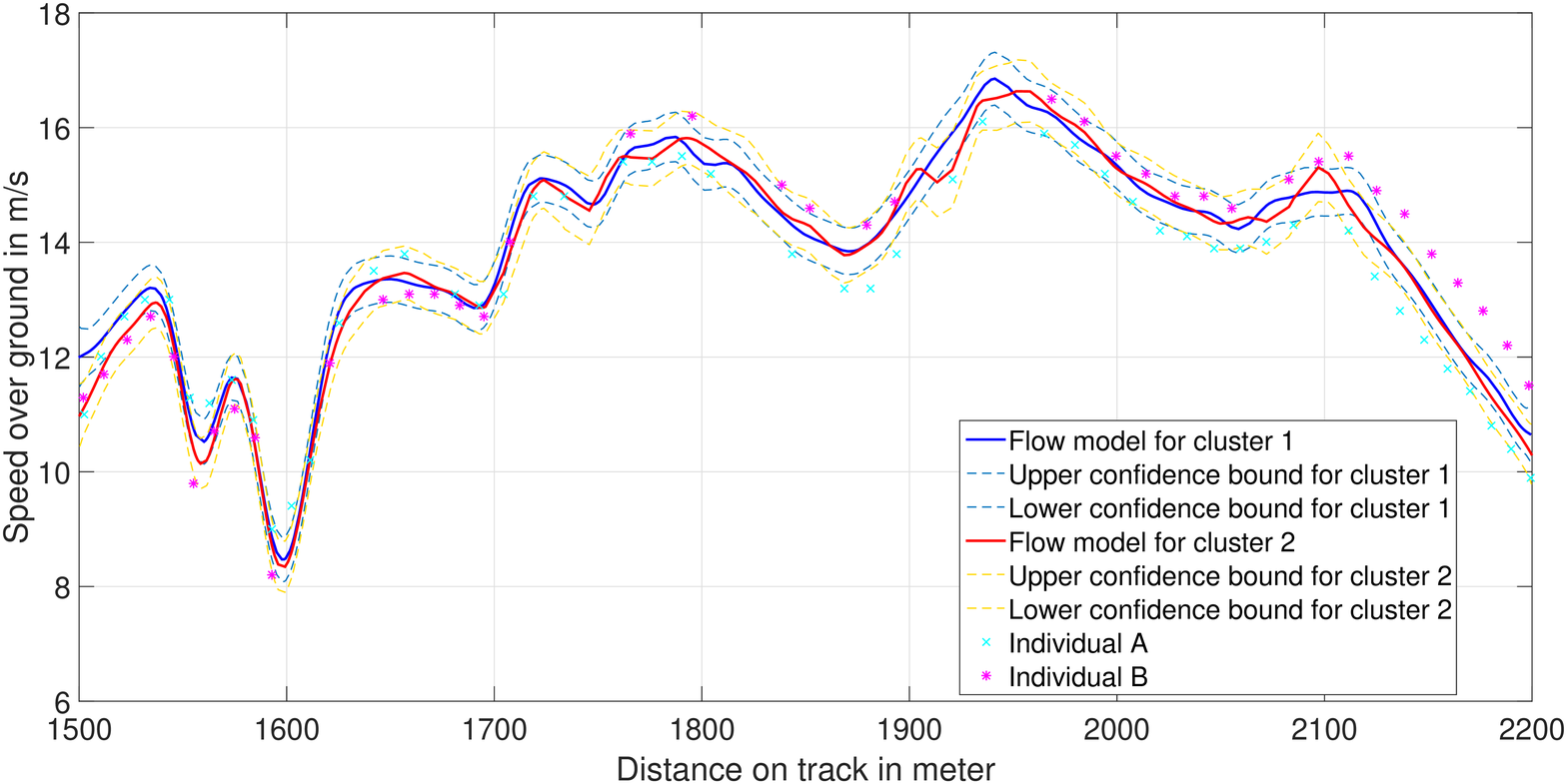}
	\caption{Flow models of individuals in lap $1$ track $1$: steepest downhill.}
	\label{fig:figure15}
\end{figure}
\begin{figure}[tb]
	\centering
	\includegraphics[width=8.5cm]{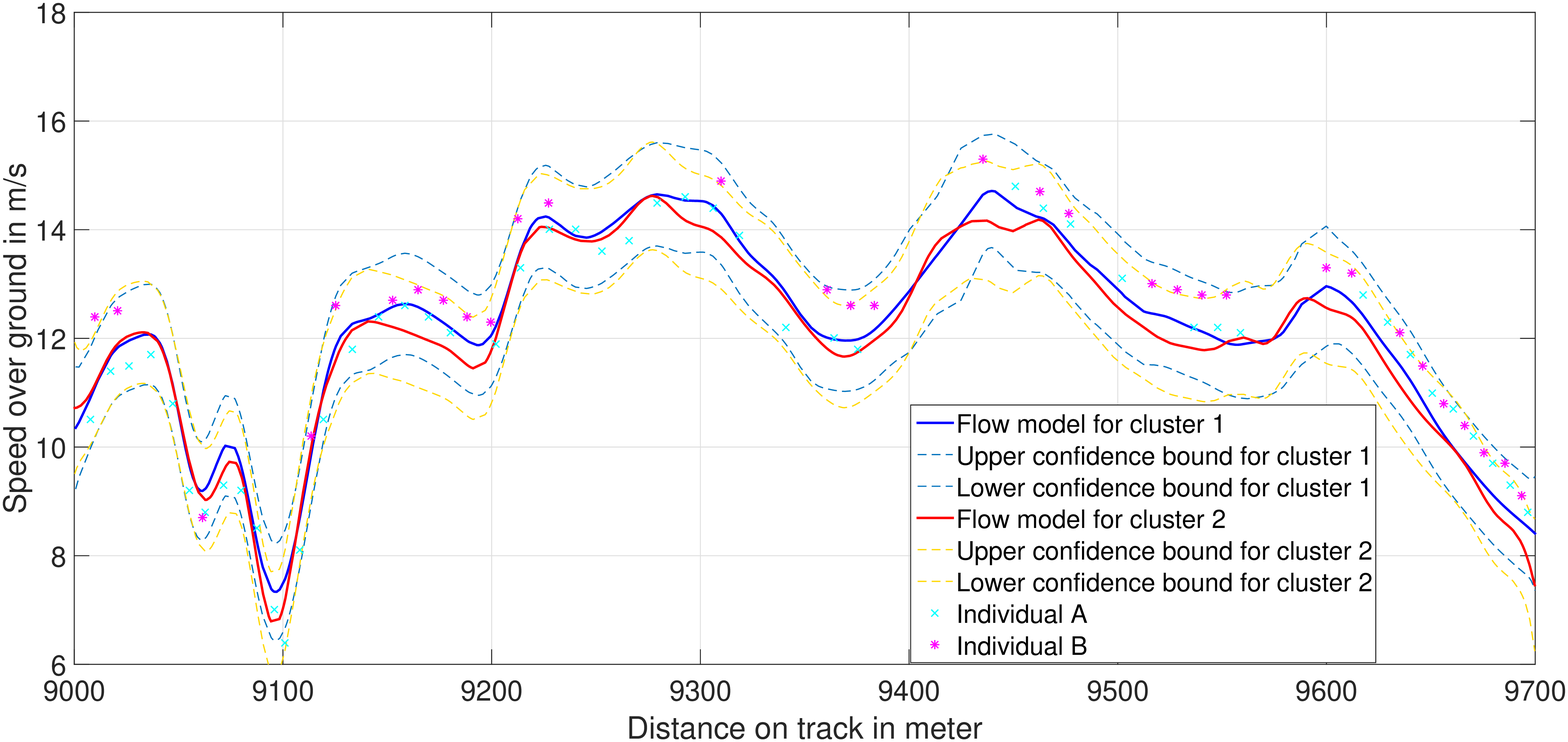}
	\caption{Flow models of individuals in lap $4$ track $1$: steepest downhill.}
	\label{fig:figure18}
\end{figure}
 After clustering, the data of individuals in the same cluster are aggregated and the GP model is applied to the aggregated data as proposed in Section \ref{subsec: FlowMulti}. The flow models for relay $1$ on track $1$ are illustrated in Fig. \ref{fig:figure11} to \ref{fig:figure18}. Due to space limitation, we only show the results for lap $1$ and $4$. Besides, the segments for individuals who perform worse (i.e., individual A) and better (i.e., individual B) in the whole competition are also plotted. In the killer hill, there is no significant difference between two clusters in lap $1$. However, the differences between two clusters become more distinct in lap $4$. This is reasonable since at the beginning of the race, all individuals may move in one cluster with similar speed. In the final stage, the difference is larger since some individuals may sprint and some may fall behind due to exhaustion. In the steepest downhill, the two clusters perform quite similarly in all laps so that one cluster is enough to model all individuals. In addition, the variance of the model becomes larger from lap $1$ to $4$. It indicates that in the steepest downhill segment, all individuals have quite uniform speed at the beginning of the race. As the race progresses, the speed of different skiers varies.  

Besides, it is observed that individuals that outperform in the killer hill have better final results in the whole competition (e.g., individual B, especially in lap $4$, has outperformed others in cluster $2$). The individuals that perform worse in the killer hill have worse final results (e.g., individual A, especially in lap $4$, has much worse performance than others in cluster $1$). This indicates that the performance in the killer hill is a more crucial factor than the steepest downhill in determining the final results.  

\begin{figure}[tb]
	\centering
	\includegraphics[width=8.5cm]{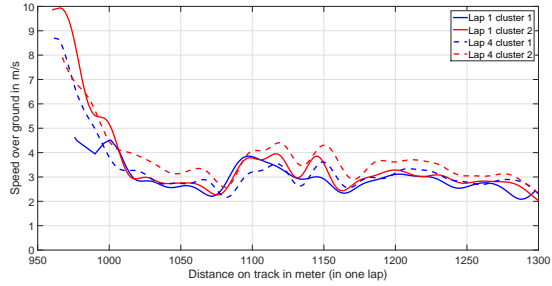}
	\caption{Flow models of individuals in lap $1$ and $4$ track $1$: killer hill.}
	\label{fig:figure19}
\end{figure}

\begin{figure}[tb]
	\centering
	\includegraphics[width=8.5cm]{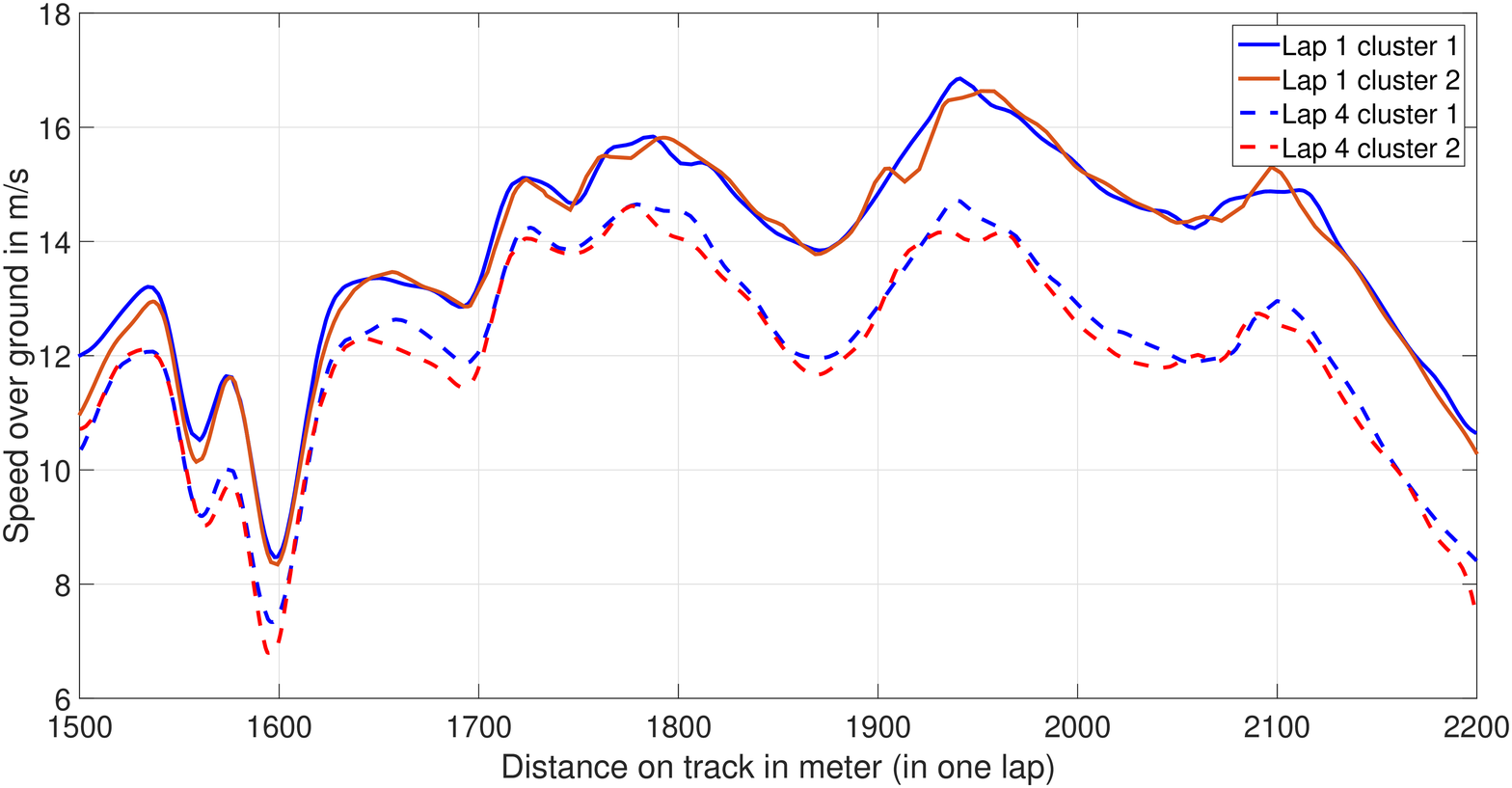}
	\caption{Flow models of individuals in lap $1$ and $4$ track $1$: steepest downhill.}
	\label{fig:figure20}
\end{figure}

Fig. \ref{fig:figure19} and \ref{fig:figure20} show a comparison between the flow models for lap $1$ and $4$ in both killer hill and steepest downhill segments. In the killer hill, all individuals almost maintain similar speed in lap $4$ as in lap $1$. However, for the steepest downhill, the average speed of all individuals are lower in lap $4$ than in lap $1$. It is probably due to different track conditions (e.g., weather condition) in lap $1$ and $4$. In lap $4$, there may be more protrusions on the track than in lap $1$ (i.e., the track is less smooth in lap $4$). Hence, the performance on the steepest downhill is greatly affected in lap $4$, while for the killer hill, the performance is mainly determined by the slope of the track.     
%Comparing the two clusters in the uphill area, quite different behaviors have been observed between $3750$ and $3850$ meters probably due to different strategies were adapted. For instance, cluster $1$ accelerate earlier and faster than cluster $2$. The variance of speed in this area for cluster $1$ is also smaller as compared to cluster $2$. However, for the two clusters in the downhill area, no significant difference in flow pattern has been observed except that cluster $2$ has higher average ground speed than cluster $1$ between $4200$ and $4500$ meters. This is probably due to the starting speed of cluster $2$ is larger. In addition, the variance for cluster $2$ is also smaller, since it is likely that individuals in cluster $2$ perform quite uniformly.   
%
\subsection{Discussions on Grey-box and Black-box Modeling}
\label{sub:dissGB}
So far we have shown the results for both the grey-box and black-box modeling approach. It is clear to see that the grey-box approach explores the partially known physical models, and the unknown part in the model are formulated by Gaussian process. In the black-box modeling approach, the model appears as random function and it can be trained based on the training inputs and outputs. We further compute and compare the predictive variance for the ground speed difference $\Delta v=v_{t+1}-v_{t}$ for both approaches. The grey-box approach yields smaller predictive variance on average. This is due to the fact that part of the model is deterministically known in grey-box approach. Hence, the ambiguity in the model is reduced. In the black-box approach, the model is completely unknown and random, which leads to larger uncertainty in prediction. 
\section{Conclusions}
\label{sec:Conclusions}
In this work, we have proposed a grey-box modeling approach for force analysis in skiing races. By analyzing the forces for different skiers, we conclude that they apply different strategies over multiple laps. For instance, skier A exerts larger forces in lap 1 and 4, where skier B exerts larger forces in lap 2 and 3, and skier C has almost evenly distributed forces for all 4 laps. Skiers having better performance are good at maintaining propulsive force at inclining area, while the declining performance is mainly determined by the friction on ice and air resistance, and larger weights lead to larger negative forces when declining. In addition, skiers having better performance are good at maintaining propulsive force at inclining area, while the declining performance is mainly determined by the friction on ice and air resistance. 

Besides, a black-box modeling approach using Gaussian process has been proposed for flow modeling. Both the standard GP and a grid based on-line GP with the local periodic kernel manifest to be powerful in modeling and predicting the performance of individuals. In particular, the grid based on-line GP reduces the computational complexity greatly while maintains similar performance. Moreover, on-line GP is more appropriate for real-time analytics where data come in sequentially. Then, clustering of individuals are performed based on the \textit{killer hill} and \textit{steepest downhill} segments. Moreover, the aggregated flow models for clusters of individuals have been developed and the results reveal that the individuals may behave differently in the killer hill, while follow a similar flow model in the steepest downhill. 

Finally, by comparing the two approaches, the grey-box approach is preferred given the real physical model is partially known, since it leads to reduced predictive variance. The black-box modeling approach, compared with the grey-box approach, is simpler and suitable when the relationship between the input and output are not clearly presented.  
\section*{Appendix A}
\label{sec:AppendixA}
\subsection*{A.1 Hyperparameters for Force Analysis}
\label{sub:HyperFA}
The ML estimates for GP in \eqref{eq:GP_resultantForce} can be obtained by maximizing the likelihood function, cf.(\ref{eq:jointDistGPFA}), with respect to $\bm{\theta}_r$, which is equivalent to
\begin{equation}
\arg \min_{\bm{\theta}_r} \, l(\bm{\theta}_r) \triangleq (\bm{F}_r - \mathbf{m}_r)^T \mathbf{C}_r^{-1} (\bm{F}_r - \mathbf{m}_r) + \ln  |\mathbf{C}_r|.
\label{eq:costFuncFA}
\end{equation} 
Various existing numerical methods can be adopted to solve this minimization problem, such as the limited-memory BFGS (LBFGS) quasi-Newton method \citep{RW06} and the conjugate gradient (CG) method. In this work, the former is adopted, which requires the first-order derivatives of the likelihood function. The first-order derivatives of $l(\bm{\theta}_r)$ can be given as 
\begin{subequations}
	\begin{align}
	\frac{\partial l(\bm{\theta}_r)}{ \partial \sigma_{n_r}^{2}} &= \textrm{tr} \! \left\lbrace \left[ \mathbf{C}_r^{-1} \!-\! \left( \mathbf{C}_r^{-1}(\mathbf{v} - \mathbf{m}) \right)\left(\cdot\right)^T \right] \! \frac{\partial \mathbf{C}_r}{\partial \sigma_{n_r}^2}  \right\rbrace   \\[0.5em] 
	\frac{\partial l(\bm{\theta}_r)}{ \partial \sigma_{r}^{2}} &= \textrm{tr} \! \left\lbrace \left[ \mathbf{C}_r^{-1} \!-\! \left( \mathbf{C}_r^{-1}(\mathbf{v} - \mathbf{m}) \right)\left(\cdot\right)^T \right] \! \frac{\partial \mathbf{C}_r}{\partial \sigma_{r}^2}  \right\rbrace   \\[0.5em]
	\frac{\partial l(\bm{\theta}_r)}{ \partial l_r} &= \textrm{tr} \! \left\lbrace \left[ \mathbf{C}_r^{-1} \!-\! \left( \mathbf{C}_r^{-1}(\mathbf{v} - \mathbf{m}) \right)\left(\cdot\right)^T \right] \! \frac{\partial \mathbf{C}_r}{\partial l_{r}} \right\rbrace   	%
%	\frac{\partial l(\bm{\theta}_r)}{ \partial l_d} &= \textrm{tr} \! \left\lbrace \left[ \mathbf{C}_r^{-1} \!-\! \left( \mathbf{C}_r^{-1}(\mathbf{v} - \mathbf{m}) \right)\left(\cdot\right)^T \right] \! \frac{\partial \mathbf{C}_r}{\partial l_d} \right\rbrace,  
	%
	%
	\end{align}   \label{eq:costFuncDeri}
\end{subequations}
where
%
%\begin{subequations}
	\begin{align}
	\frac{\partial \mathbf{C}_r}{\partial \sigma_{n_r}^2} &= \mathbf{I}_{N}. \nonumber  \\
	\left[ \frac{\partial \mathbf{C}_r}{\partial \sigma_{r}^2} \right]_{j,k} \!\!\!\! &= \begin{cases}
	1, & \!\! j = k  \\[0.5em]
	\exp \! \left[ \frac{-(d_{j}-d_{k})^2}{l_{r}^2} \right] , & \!\! j \neq k
	\end{cases}  \nonumber \\
	\left[ \frac{\partial \mathbf{C}_r}{\partial l_{r}} \right]_{j,k} \!\!\!\! &= \begin{cases}
	0, & \!\! j=k,  \\
	2\sigma_{r}^2 \! \exp \! \left[ \frac{-(d_{j} -d_{k})^2}{l_{r}^2} \right] \! \frac{(d_{j}-d_{k})^2}{l_{r}^3}  , & \!\! j \neq k.
	\end{cases}   \nonumber
	\end{align}
%\end{subequations}
Here we use $(A)(\cdot)^T$ to denote $(A)(A)^T$ for brevity.
\subsection*{A.2 Hyperparameters for Individual Model}
\label{sub:HyperIndi}
The maximum-likelihood estimate of the GPR model parameters, $\hat{\bm{\theta}}$, can be obtained by maximizing the Gaussian prior likelihood function similarly.% with respect to $\bm{\theta}$, which is equivalent to 
%%
%\begin{equation}
%\arg \min_{\bm{\theta}} \, l(\bm{\theta}) \triangleq (\mathbf{v} - \mathbf{m})^T \mathbf{C}^{-1} (\mathbf{v} - \mathbf{m}) + \ln  |\mathbf{C}|.
%\label{eq:costFuncLBFGS}
%\end{equation} 
%
 The first-order derivatives can be given as the same form as in \eqref{eq:costFuncDeri}, except that
%
%
%\begin{subequations}
%	%	
%	\begin{align}
%	%
%	\frac{\partial l(\bm{\theta})}{ \partial \sigma_{n}^{2}} &= \textrm{tr} \! \left\lbrace \left[ \mathbf{C}^{-1} \!-\! \left( \mathbf{C}^{-1}(\mathbf{v} - \mathbf{m}) \right)\left(\cdot\right)^T \right] \! \frac{\partial \mathbf{C}}{\partial \sigma_{n}^2}  \right\rbrace   \\[0.5em] 
%	%
%	\frac{\partial l(\bm{\theta})}{ \partial \sigma_{s}^{2}} &= \textrm{tr} \! \left\lbrace \left[ \mathbf{C}^{-1} \!-\! \left( \mathbf{C}^{-1}(\mathbf{v} - \mathbf{m}) \right)\left(\cdot\right)^T \right] \! \frac{\partial \mathbf{C}}{\partial \sigma_{s}^2}  \right\rbrace   \\[0.5em]
%	%
%	\frac{\partial l(\bm{\theta})}{ \partial l_p} &= \textrm{tr} \! \left\lbrace \left[ \mathbf{C}^{-1} \!-\! \left( \mathbf{C}^{-1}(\mathbf{v} - \mathbf{m}) \right)\left(\cdot\right)^T \right] \! \frac{\partial \mathbf{C}}{\partial l_{p}} \right\rbrace   \\[0.5em]
%	%
%	\frac{\partial l(\bm{\theta})}{ \partial l_d} &= \textrm{tr} \! \left\lbrace \left[ \mathbf{C}^{-1} \!-\! \left( \mathbf{C}^{-1}(\mathbf{v} - \mathbf{m}) \right)\left(\cdot\right)^T \right] \! \frac{\partial \mathbf{C}}{\partial l_d} \right\rbrace,  
%	%
%	%
%	\end{align}   \label{eq:costFuncDeri}
%	%
%\end{subequations}
%
%
%
%\begin{subequations}
%
\begin{align}
\frac{\partial \mathbf{C}}{\partial \sigma_{n}^2} &= \mathbf{I}_{M}  \nonumber \\
\left[ \frac{\partial \mathbf{C}}{\partial \sigma_{s}^2} \right]_{j,k} \!\!\!\! &= \begin{cases}
1, & \!\! j = k \nonumber \\[0.5em]
\exp \! \left[ \frac{-\sin^2\left(\frac{(d_{j}-d_{k})\pi}{\lambda}\right)}{l_{p}^2} \right] \! \exp \! \left[ \frac{-(d_{j}-d_{k})^2}{l_{d}^2} \right] , & \!\! j \neq k
\end{cases}  \nonumber \\[0.5em]
\left[ \frac{\partial \mathbf{C}}{\partial l_{p}} \right]_{j,k} \!\!\!\! &= \begin{cases}
0, & \!\! j=k \nonumber \\
2k(d_j, d_k) \! \sin^2\left(\frac{(d_{j}-d_{k})\pi}{\lambda}\right) \! \frac{1}{l_{p}^3}, & \!\! j \neq k
\end{cases}  \nonumber \\[0.5em]
\left[ \frac{\partial \mathbf{C}}{\partial l_{d}} \right]_{j,k} \!\!\!\! &= \begin{cases}
0, & \!\! j=k \nonumber \\
2k(d_j, d_k) \! \frac{(d_{j}-d_{k})^2}{l_{d}^3}  , & \!\! j \neq k. \nonumber 
\end{cases}  
\end{align}
%
%\end{subequations}
%
\subsection*{A.3 Hyperparameters for Aggregated Model}
\label{sub:HyperMul}
Similarly, the ML estimate for the multiple skiers GP model parameters can be obtained by maximizing the likelihood function, cf.(\ref{eq:GPdistributionCluster}), with respect to $\bm{\theta}_c$.%, which is equivalent to
%\begin{equation}
%\arg \min_{\bm{\theta}_c} \, l(\bm{\theta}_c) \triangleq (\mathbf{v} - \mathbf{m})^T \mathbf{C}_c^{-1} (\mathbf{v} - \mathbf{m}_c) + \ln  |\mathbf{C}_c|.
%\label{eq:costFuncCluster}
%\end{equation} 
The first-order derivatives of the cost function, $l_c(\bm{\theta}_c)$, have similar formats as given in \eqref{eq:costFuncDeri}, except that 
%\begin{subequations}
%		
%	\begin{align}
%	%
%	\frac{\partial l(\bm{\theta}_c)}{ \partial \sigma_{n}^{2}} &= \textrm{tr} \! \left\lbrace \left[ \mathbf{C}_c^{-1} \!-\! \left( \mathbf{C}_c^{-1}(\mathbf{v} - \mathbf{m}) \right)\left(\cdot\right)^T \right] \! \frac{\partial \mathbf{C}_c}{\partial \sigma_{n}^2}  \right\rbrace   \\[0.5em]
%	%
%	\frac{\partial l(\bm{\theta}_c)}{ \partial \sigma_{s}^{2}} &= \textrm{tr} \! \left\lbrace \left[ \mathbf{C}_c^{-1} \!-\! \left( \mathbf{C}_c^{-1}(\mathbf{v} - \mathbf{m}) \right)\left(\cdot\right)^T \right] \! \frac{\partial \mathbf{C}_c}{\partial \sigma_{s}^2}  \right\rbrace   \\[0.5em]
%	%
%	\frac{\partial l(\bm{\theta}_c)}{ \partial l_d} &= \textrm{tr} \! \left\lbrace \left[ \mathbf{C}_c^{-1} \!-\! \left( \mathbf{C}_c^{-1}(\mathbf{v} - \mathbf{m}) \right)\left(\cdot\right)^T \right] \! \frac{\partial \mathbf{C}_c}{\partial l_d} \right\rbrace   \\[0.5em]
%	%
%	\frac{\partial l(\bm{\theta}_c)}{ \partial \sigma_{c}^{2}} &= \textrm{tr} \! \left\lbrace \left[ \mathbf{C}_c^{-1} \!-\! \left( \mathbf{C}_c^{-1}(\mathbf{v} - \mathbf{m}) \right)\left(\cdot\right)^T \right] \! \frac{\partial \mathbf{C}_c}{\partial \sigma_{c}^2}  \right\rbrace   \\[0.5em]
%	%
%	\frac{\partial l(\bm{\theta}_c)}{ \partial l_c} &= \textrm{tr} \! \left\lbrace \left[ \mathbf{C}_c^{-1} \!-\! \left( \mathbf{C}_c^{-1}(\mathbf{v} - \mathbf{m}) \right)\left(\cdot\right)^T \right] \! \frac{\partial \mathbf{C}_c}{\partial l_c} \right\rbrace,   
%	%
%	\end{align}   
%	
%\end{subequations}
%
%where 

\begin{subequations}
	\begin{align}
	\frac{\partial \mathbf{C}_c}{\partial \sigma_{n}^2} &= \mathbf{I}_{N_D}.  \nonumber \\
	\left[ \frac{\partial \mathbf{C}_c}{\partial \sigma_{s}^2} \right]_{j,k} \!\!\!\! &= \begin{cases}
	1, & \!\! j = k \nonumber \\[0.5em]
	\exp \! \left[ \frac{-(d_{j}-d_{k})^2}{l_{d}^2} \right] , & \!\! j \neq k
	\end{cases}  \nonumber \\[0.5em]
	\left[ \frac{\partial \mathbf{C}_c}{\partial l_{d}} \right]_{j,k} \!\!\!\! &= \begin{cases}
	0, & \!\! j=k \nonumber \\
	2\sigma_{s}^2 \! \exp \! \left[ \frac{-(d_{j} -d_{k})^2}{l_{d}^2} \right] \! \frac{(d_{j}-d_{k})^2}{l_{d}^3}  , & \!\! j \neq k
	\end{cases}  \nonumber \\[0.5em]
	\left[ \frac{\partial \mathbf{C}_c}{\partial \sigma_{c}^2} \right]_{j,k} \!\!\!\! &= \begin{cases}
	1, & \!\! j = k \nonumber \\[0.5em]
	\exp \! \left[ \frac{-(d_{j}-d_{k})^2}{l_{c}^2} \right] , & \!\! j \neq k
	\end{cases}  \nonumber \\[0.5em]
	\left[ \frac{\partial \mathbf{C}_c}{\partial l_{c}} \right]_{j,k} \!\!\!\! &= \begin{cases}
	0, & \!\! j=k \nonumber \\
	2\sigma_{c}^2 \! \exp \! \left[ \frac{-(d_{j} -d_{k})^2}{l_{c}^2} \right] \! \frac{(d_{j}-d_{k})^2}{l_{c}^3}  , & \!\! j \neq k \nonumber. 
	\end{cases}  
	\end{align}
\end{subequations}

%
%\left[ \frac{\partial \mathbf{C}_c}{\partial \sigma_{c}^2} \right]_{j,k} \!\!\!\! &= \begin{cases}
%1, & \!\! j = k \nonumber \\[0.5em]
%\exp \! \left[ \frac{-(d_{j}-d_{k})^2}{l_{c}^2} \right] , & \!\! j \neq k
%\end{cases}  \nonumber \\[0.5em]
%%
%\left[ \frac{\partial \mathbf{C}_c}{\partial l_{c}} \right]_{j,k} \!\!\!\! &= \begin{cases}
%0, & \!\! j=k \nonumber \\
%2\sigma_{c}^2 \! \exp \! \left[ \frac{-(d_{j} -d_{k})^2}{l_{c}^2} \right] \! \frac{(d_{j}-d_{k})^2}{l_{c}^3}  , & \!\! j \neq k \nonumber 
%\end{cases}  
%

\section*{Appendix B}
\label{sec:AppendixB}
Imagine that $\mathcal{S}_g \triangleq \{\bar{\mathbf{v}}, \bar{\mathbf{d}}\}$ is also a training dataset despite that $\bar{\mathbf{v}}(\bar{\mathbf{d}})$ is latent. Given a novel input, $d_{*}$, the posterior distribution of observing a noisy $v(d_{*})$, given $\mathcal{S}_{g}$, can be easily obtained as follows: 
\begin{equation}
p(v(d_{*}) | \mathcal{S}_{g}) \sim \mathcal{N} \left( \mu_{g}^{p}, \sigma_{g}^{2, p} \right), 
\end{equation}
where 
\begin{subequations}
	\begin{align}
	\mu_{g}^{p} &= \mathbf{k}(d_{*}, \bar{\mathbf{d}})^T \bar{\mathbf{K}}^{-1} (\bar{\mathbf{v}} - \bar{\mathbf{m}}) + m(d_{*}) \\
	\sigma_{g}^{2, p} &= \sigma_{s}^{2}  + \sigma_{n}^{2} - \mathbf{k}(d_{*}, \bar{\mathbf{d}})^T \bar{\mathbf{K}}^{-1} \mathbf{k}(d_{*}, \bar{\mathbf{d}}).
	\end{align}
\end{subequations}	
The posterior distribution of $v(d_{*})$, given $\mathcal{S}$ and $\bar{\mathbf{d}}$, can be computed analytically via the following marginalization:
\begin{equation}
p( v(d_{*}) | \mathcal{S}, \bar{\mathbf{d}} ) \!=\!\! \int \! p(\bar{\mathbf{v}} | \bar{\mathbf{d}}, \mathcal{S})  p( v(d_{*}) | \mathcal{S}_{g}, \mathcal{S})  \rm{d} \bar{\mathbf{v}} , \nonumber 
\end{equation}
and approximated with reduced computational complexity, like in \citep{SG06}, by 
\begin{equation}
p( v(d_{*}) | \mathcal{S}, \bar{\mathbf{d}} ) \!\approx\!\!\! \int \! \!p(\bar{\mathbf{v}} | \bar{\mathbf{d}}, \mathcal{S})  p( v(d_{*}) | \mathcal{S}_{g})  \rm{d} \bar{\mathbf{v}}. \nonumber 
\end{equation}
Since both $p(v(d_{*}) | \mathcal{S}_{g})$ and $p(\bar{\mathbf{v}} | \bar{\mathbf{d}}, \mathcal{S})$ are Gaussian distributed, applying Lemma A.1 in \citep{Sarkka13} yields eventually (\ref{eq:mu-onlineGP}) and (\ref{eq:var-onlineGP}).

%\begin{equation}
%\end{equation}

%\begin{figure}
%% Use the relevant command to insert your figure file.
%% For example, with the graphicx package use
%  \includegraphics{example.eps}
%\end{figure}
%\begin{figure*}
%% Use the relevant command to insert your figure file.
%% For example, with the graphicx package use
%  \includegraphics[width=0.75\textwidth]{example.eps}
%\end{figure*}
%\begin{table}
%\begin{tabular}{lll}
%\hline\noalign{\smallskip}
%\end{tabular}
%\end{table}

\begin{acknowledgements}
\label{sec:ack}
This work is funded by European Union FP7 Marie Curie training programme on Tracking in Complex Sensor Systems (TRAX) with grant number 607400 from 2014 to 2017. This work is also funded by ELLIT project, which is a strategic research environment funded by the Swedish government in 2010. Feng Yin is mainly funded by Shenzhen Science and Technology Innovation Council under Grant JCYJ20170307155957688, Guangdong Province Pearl River Talent Team under Grant 2017ZT07X152, and partly by the Shenzhen Fundamental Research Fund under Grant (Key Lab) ZDSYS201707251409055. 
\end{acknowledgements}


\bibliographystyle{plainnat}
\bibliography{Main}   % name your BibTeX data base

%\begin{thebibliography}{}
%%
%% and use \bibitem to create references. Consult the Instructions
%% for authors for reference list style.
%\end{thebibliography}

\end{document}